\documentclass[sigconf,screen,nonacm]{acmart}

\usepackage{algorithm}
\usepackage{algorithmic}
\usepackage{multirow}
\usepackage{colortbl}
\usepackage{cleveref}
\usepackage{balance}

\AtBeginDocument{%
  }

\begin{document}

\title{Contrastive Regularization over LoRA for Multimodal Biomedical Image Incremental Learning}

\author{Haojie Zhang}
\orcid{0009-0000-4223-7717}
\email{234711081@csu.edu.cn}
\affiliation{%
  \institution{School of Computer Science and Engineering, Central South University}
  \city{Changsha}
  \state{Hunan}
  \country{China}
}

\author{Yixiong Liang}
\orcid{0000-0003-0407-5838}
\email{yxliang@csu.edu.cn}
\affiliation{%
  \institution{School of Computer Science and Engineering, Central South University}
  \city{Changsha}
  \state{Hunan}
  \country{China}
}

\author{Hulin Kuang}
\orcid{0000-0001-7341-9871}
\email{hulinkuang@csu.edu.cn}
\affiliation{%
  \institution{School of Computer Science and Engineering, Central South University}
  \city{Changsha}
  \state{Hunan}
  \country{China}
}

\author{Lihui Cen}
\orcid{0000-0002-5323-4818}
\email{lhcen@csu.edu.cn}
\affiliation{%
  \institution{School of Automation, Central South University}
  \city{Changsha}
  \state{Hunan}
  \country{China}
}

\author{Zhe Qu}
\orcid{0000-0003-2211-2137}
\email{zhe_qu@csu.edu.cn}
\affiliation{%
  \institution{School of Computer Science and Engineering, Central South University}
  \city{Changsha}
  \state{Hunan}
  \country{China}
}

\author{Yigang Cen}
\orcid{0000-0001-6255-9422}
\email{ygcen@bjtu.edu.cn}
\affiliation{%
  \institution{School of Computer Science and Technology, Beijing Jiaotong University}
  \city{Beijing}
  \state{Beijing}
  \country{China}
}

\author{Min Zeng}
\orcid{0000-0002-1726-0955}
\email{zengmin@csu.edu.cn}
\affiliation{%
  \institution{School of Computer Science and Engineering, Central South University}
  \city{Changsha}
  \state{Hunan}
  \country{China}
}

\author{Shichao Kan}
\orcid{0000-0003-0097-6196}
\email{kanshichao@csu.edu.cn}
\authornote{Corresponding author}
\affiliation{%
  \institution{School of Computer Science and Engineering, Central South University}
  \city{Changsha}
  \state{Hunan}
  \country{China}
}

\renewcommand{\shortauthors}{Haojie Zhang, et al.}

\begin{abstract}
Multimodal Biomedical Image Incremental Learning (MBIIL) is essential for handling diverse tasks and modalities in the biomedical domain, as training separate models for each modality or task significantly increases inference costs. 
Existing incremental learning methods focus on task expansion within a single modality, whereas MBIIL seeks to train a unified model incrementally across modalities. 
The MBIIL faces two challenges: \textbf{\textit{I) How to preserve previously learned knowledge during incremental updates? II) How to effectively leverage knowledge acquired from existing modalities to support new modalities?}}
To address these challenges, we propose MSLoRA-CR, a method that fine-tunes \textbf{M}odality-\textbf{S}pecific \textbf{LoRA} modules while incorporating \textbf{C}ontrastive \textbf{R}egularization to enhance intra-modality knowledge sharing and promote inter-modality knowledge differentiation. 
Our approach builds upon a large vision-language model (LVLM), keeping the pretrained model frozen while incrementally adapting new LoRA modules for each modality or task.
Experiments on the incremental learning of biomedical images demonstrate that MSLoRA-CR outperforms both the state-of-the-art (SOTA) approach of training separate models for each modality and the general incremental learning method (incrementally fine-tuning LoRA). 
Specifically, MSLoRA-CR achieves a 1.88\% improvement in overall performance compared to unconstrained incremental learning methods while maintaining computational efficiency. 
\textit{Our code is publicly available at \url{https://github.com/VentusAislant/MSLoRA_CR}.}
\end{abstract}

\keywords{Contrastive Regularization, Biomedical Image Incremental Learning}

\maketitle

\vspace{-2mm}
\section{Introduction}
\begin{figure}[t]
    \centering
    \includegraphics[width=1\linewidth]{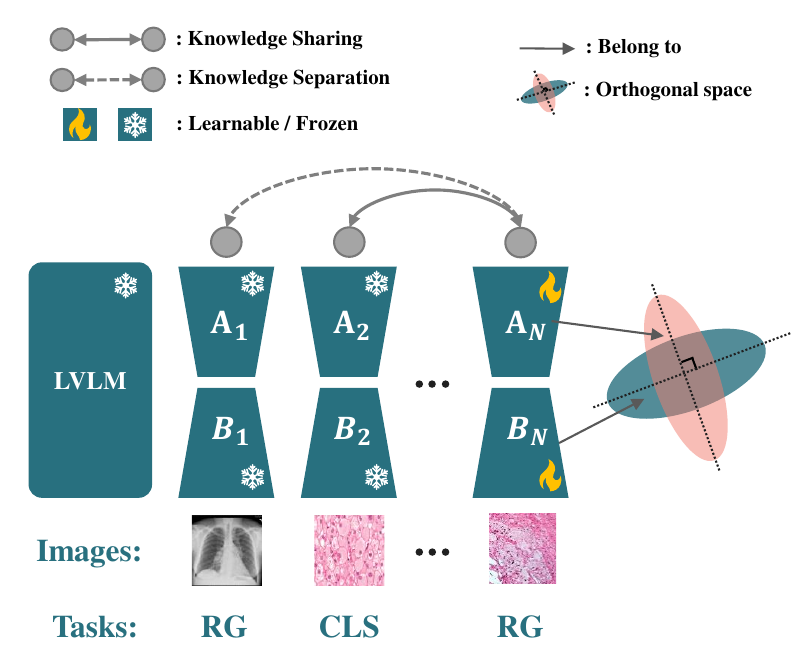}
    \caption{Contrastive regularization aligns LoRA parameters within the same modality for knowledge sharing while separating different modalities to reduce interference. Moreover, an orthogonality constraint between A and B further reinforces this process. `RG` and `CLS` denote Report Generation and Classification task, respectively.}
    \label{fig:intro}
  \vspace{-2mm}
\end{figure}

Incremental learning requires models to sequentially learn multiple tasks, balancing two key aspects: maintaining performance on previously learned tasks (stability) and adapting to new ones (plasticity) \cite{intro-survey-cl, intro-cl}. While most incremental learning algorithms focus on task increments \cite{intro-task-il, intro-task-il2-packnet, intro-task-il3} within a single modality, the biomedical domain presents a unique challenge due to the diversity of image modalities, such as radiology and pathology. This complexity increases the difficulty of ensuring stability and plasticity. To address this, we introduce the MBIIL task, aimed at incrementally training a unified model that can handle various biomedical image modalities and tasks. The MBIIL task presents two major challenges: preserving performance across previously learned modalities and tasks during updates, and efficiently utilizing pretrained models to support learning of new tasks.

Recent advances in parameter-efficient fine-tuning (PEFT) methods \cite{intro-peft, intro-peft2, intro-preft3, intro-peft4-tip-adapter, hydra-lora, hira} have facilitated incremental learning, with LoRA \cite{lora} attracting significant attention.
These methods involve freezing a pretrained model while fine-tuning a small set of learnable parameters for each new task. While this approach addresses the challenge of task adaptation, it does not account for interactions between new and existing LoRA parameters, which can lead to suboptimal performance. To overcome this and tackle the second challenge of knowledge sharing, InfLoRA \cite{inflora} introduces a subspace matrix to prevent interference between new and previously learned tasks. The mixture of experts (MoE) \cite{intro-moe} adapter further enhances this by learning task-specific routing to adaptively merge LoRA parameters, particularly in the CLIP \cite{intro-clip} framework. However, these methods are designed primarily for task-incremental learning with natural scene images, and do not consider the complexities of modality-incremental learning.

In this paper, we address the MBIIL problem by applying task-incremental learning across diverse biomedical image modalities. Since task-incremental learning provides task identities during inference, the model can maintain performance on previously learned tasks while adapting to new ones. We adopt this strategy to fine-tune the LoRA parameters incrementally for new tasks, thereby preserving the performance of the learned modalities and tasks. 
Specifically, we introduce modality-specific LoRA modules to accommodate new tasks in a sequential learning setup. When a new task arrives, we dynamically expand a new LoRA branch tailored to the current task, with a theoretically proven balance between plasticity and stability in MBIIL.

To enhance knowledge utilization in this incremental process, we introduce a contrastive regularization mechanism that strategically structures LoRA parameter relationships. As shown in \cref{fig:intro}, \textbf{this mechanism pulls the LoRA parameters of the current task closer to those of previously learned tasks within the same modality, facilitating knowledge sharing. Simultaneously, it pushes them apart from those of different modalities to prevent knowledge interference.} Moreover, we incorporate an orthogonality constraint between key components to reinforce modality separation and improve learning efficiency. This design ensures that newly introduced tasks effectively leverage prior knowledge while maintaining distinct representations for different modalities.

The key contributions of this work are as follows:
\begin{itemize}
     \item We introduce the MBIIL problem to address the challenge of large modality gaps in the continual learning of large vision-language models for biomedical applications. To address this, we designed the MSLoRA-CR module, which balances plasticity and stability in the MBIIL process.
     \item We develop a contrastive regularization loss to facilitate knowledge sharing within the same modality while mitigating knowledge conflicts across different modalities.
     \item We conducted extensive experiments for the MBIIL task and the results demonstrate the effectiveness of our approach, achieving a 1.88\% improvement in overall performance while maintaining high efficiency.
\end{itemize}

\section{Related Works}
\label{sec:related works}


\paragraph{\textbf{Continual Learning}}

Continual learning enables models to incrementally learn new tasks while preserving prior knowledge, addressing catastrophic forgetting by balancing plasticity and stability.  
Mainstream approaches fall into three categories: \textit{regularization-based}, \textit{replay-based}, and \textit{architecture-based}.  
\textit{Regularization-based} methods \cite{cl-r-ewc, cl-r-is, cl-r-mas, cl-r-GPM, jiang2025unlocking} incorporate an additional term into the loss function, constraining weight updates to preserve previously learned knowledge by minimizing significant deviations in parameters critical to past tasks.
\textit{Replay-based} methods \cite{cl-m-replay, cl-m-gem, cl-m-icarl} use memory buffers to replay past data during learning of new tasks, mitigating forgetting.
\textit{Architecture-based} methods \cite{cl-a-expert-gate, cl-a-DNE, cl-a-dytox} partition the model into a shared component for task-agnostic knowledge and dynamic task-specific modules that expand as new tasks emerge, enabling continual adaptation. 
Our work can be categorized into the architecture-based method, which continually learns the modality-specific LoRA in an incremental learning manner.

\paragraph{\textbf{Parameter Efficient Fine-Tuning}}
As model sizes grow, full fine-tuning becomes costly and prone to overfitting on small datasets. Parameter-Efficient Fine-Tuning (PEFT) methods adapt downstream tasks by freezing pretrained models and injecting few trainable parameters. 
Mainstream PEFT methods can be categorized into three types: \textit{adapter-based}, \textit{prompt-based}, and \textit{delta-based} approaches.
\textit{Adapter-based} methods \cite{adapter, clip-adapter} integrate additional trainable components, such as linear layers, into pretrained models, enabling adaptation to downstream tasks while preserving the original model's parameters.
\textit{Prompt-based} methods \cite{prompt, coop, cocoop} avoid adding extra modules by prepending a learnable embedding matrix to the input embeddings of pretrained attention models, guiding their adaptation to downstream tasks.
\textit{Delta-based} methods, such as LoRA \cite{lora, adalora, hydra-lora, hira}, inject low-rank learnable delta parameters into specific layers without changing the model architecture; these parameters can be merged back after training. Despite having fewer trainable parameters, delta-based methods match or exceed full fine-tuning performance, as parameter training occurs within a constrained subspace.

With the rise of pretrained models and PEFT techniques, continual learning methods leveraging PEFT approaches have gained traction \cite{cl-peft-ddas, inflora, cl-peft-CIL, sd-lora}. Pretrained models effectively preserve task-agnostic knowledge, while architecture-based PEFT methods facilitate adaptation to new tasks by introducing minimal parameters and minimally disrupting the pretrained model. Our work is also developed based on the LoRA fine-tuning.

\paragraph{\textbf{Biomedical Large Vision Language Model Training}}
Biomedical large vision-language models adapt general LVLMs \cite{flamingo, blip2, llava, minigpt4} for biomedical applications, where diverse modalities and complex tasks pose greater challenges than natural images. Existing methods mainly follow two routes: \textit{joint training} and \textit{pretraining with task-specific fine-tuning}.
To train more generalized and practical biomedical large models, existing research has explored two main approaches: \textit{joint training} methods and \textit{pretraining with task-specific fine-tuning}.
\textit{Joint training} \cite{llava-med, BiomedGPT, GBAI} integrates multiple modalities/tasks into a unified format for full fine-tuning to create multifunctional models. However, it requires heavy computation and couples modalities and tasks, which reduces interpretability.
On the other hand,\textit{Pretraining with task-specific fine-tuning} \cite{pathasst, xraychat, skingpt} first pretrains on general biomedical knowledge, then fine-tunes for specific tasks. This paradigm decouples tasks but requires training separate models, leading to inference inefficiency, parameter redundancy, and neglect of relationships among biomedical modalities.

In contrast, we frame building a general medical vision-language model as a continual learning problem, leveraging continual learning to advance biomedical AGI. This enables end-to-end training that reduces computational and time costs compared to joint training, while effectively decoupling tasks and modalities, improving interpretability and reliability—key for medical applications.

\vspace{-1mm}
\section{Methodology}
\label{sec:methods}
We first give the definition of the proposed MBIIL problem. Given a set of tasks $\mathcal{A}:=\{a^n\}_{n=1}^A$ and a set of biomedical modalities $\mathcal{B}:=\{b^n\}_{n=1}^B$.
There is a modality-specific task sequence to be learned $\mathcal{T} := \{(a_i, b_j) \mid a_i \in \mathcal{A}, b_j \in \mathcal{B}, \; 1 \leq i \leq |\mathcal{A}|, \; 1 \leq j \leq |\mathcal{B}|\}$.
The modality-specific task dataset $\mathcal{D}_t:=\{(\mathcal{X}_i, \mathcal{Y}_i)\}_{i=1}^T$, where $\mathcal{X}_i$ represents the input of samples in training set and $\mathcal{Y}_i$ is the corresponding label.
The goal of the MBIIL problem is to enable a model to continually learn task sequences without forgetting previous knowledge, while also mitigating performance degradation from biomedical modality conflicts. This work builds on Large Vision Language Models (LVLMs) and Low-Rank Adaptation (LoRA); we briefly introduce them as preliminaries below.

\subsection{Preliminaries}

\paragraph{\textbf{Large Vision Language Model}}
We adopt an LLaVA-like large vision-language model as our base architecture. Specifically, it consists of three components: a visual encoder, denoted as $g_{\theta_v}(\cdot)$, a mapping layer, denoted as $h_{\theta_m}(\cdot)$, and a large language model, denoted as $f_{\theta_t}(\cdot)$.
Given an image $\mathbf{X}_v \in \mathbb{R}^{C \times H \times V}$, where $C$ represents the number of channels, and $H$ and $V$ represent the image resolution, we first apply the visual encoder (CLIP ViT) to extract features, $\mathbf{e}_v = g_{\theta_v}(\mathbf{X}_v) \in \mathbb{R}^{P \times d_v}$, where $P$ represents the number of patches and $d_v$ represents the hidden dimension of ViT. Then, the mapping layer is used to project the visual representation into the feature space understood by the large language model, $\mathbf{e}_v' = h_{\theta_m}(\mathbf{e}_v) \in \mathbb{R}^{P \times d_t}$, where $d_t$ is the hidden dimension of the language model. Finally, the visual features and text instruction features $\mathbf{e}_t \in \mathbb{R}^{L \times d_t}$ are concatenated and fed into the large language model, yielding $\mathbf{e}_o = f_{\theta_t}(\mathbf{e}_v'; \mathbf{e}_t)$.

\paragraph{\textbf{Low-Rank Adaption}}
For incremental learning with LVLM, LoRA is one of the most popular PEFT methods. It introduces a trainable weight $\mathbf{\mathbf{\Delta W}}$ for each linear layer with weight $\mathbf{W} \in \mathbb{R}^{d_o \times d_i}$, where $d_o$ represents the output dimension and $d_i$ represents the input dimension. This weight is expressed using low-rank decomposition, where $\mathbf{\Delta W}$ is the product of two low-rank matrices, $\mathbf{A} \in \mathbb{R}^{d_o \times r}$ and $\mathbf{B} \in \mathbb{R}^{r \times d_i}$. The rank $r$ is much smaller than both $d_o$ and $d_i$, and $\mathbf{\Delta W} = \mathbf{A}\mathbf{B}$. Assuming the inputs and outputs of the linear layer are $\mathbf{h}$ and $\mathbf{e}$, respectively, then $\mathbf{e} = (\mathbf{W} + \mathbf{\Delta W})\mathbf{h} = \mathbf{W}\mathbf{h} + \mathbf{A}\mathbf{B}\mathbf{h}$. In the initialization setup of LoRA, $\mathbf{A}$ is initialized with zeros, and $\mathbf{B}$ is initialized using a normal distribution.

In the following subsections, we provide the framework overview of our proposed MSLoRA-CR method for MBIIL problem, and given details about modality specific LoRA fine-tuning and contrastive regularization loss.

\begin{figure*}[ht]
    \centering
    \includegraphics[width=1\linewidth]{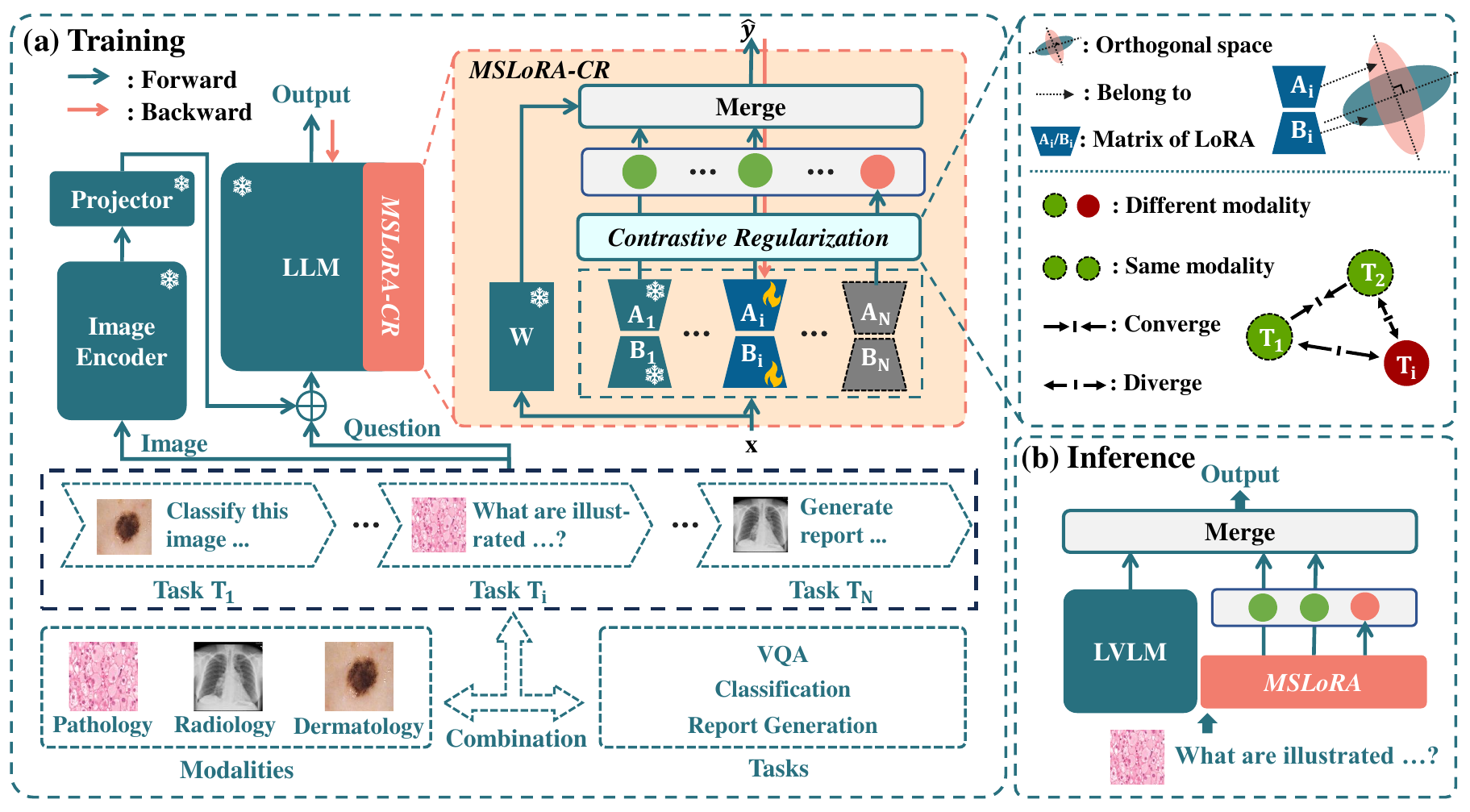}
    \caption{The overview of modality-specific LoRA fine-tuning with contrastive regularization (MSLoRA-CR).}
    \label{fig:overview}
\end{figure*}

\subsection{Framework Overview}

The overall framework of our proposed method is shown in \cref{fig:overview}. First, different tasks and modalities are combined to form a modality-specific task sequence. 
To enable continual learning in this task sequence, we keep the entire large vision-language model frozen, including the visual encoder, mapping layer, and language model, while integrating trainable MSLoRA-CR modules within the LLM.
The MSLoRA-CR module dynamically expands with each new task, learning task-specific parameters while preserving previously learned knowledge. In this module, we utilize a Contrastive Regularization (CR) loss to integrate past knowledge and regularize the parameter subspace for new tasks.

During inference, a masking strategy is applied to merge task- and modality-specific LoRA parameters, maintaining model performance across all tasks and modalities.

\subsection{Modality Specific LoRA Fine-tuning}
In the following, we introduce incremental learning with MSLoRA method and provide detailed theoretical analysis.

\paragraph{\textbf{Method Introduction}}
We propose the Modality-Specific LoRA (MSLoRA) fine-tuning module, which utilizes multiple LoRA branches to learn distinct tasks, with task-specific masks controlling their merging. Specifically, when the $t$-th task arrives, a new LoRA branch is added to learn this task, while the existing $t-1$ branches retain information from previous tasks. A mask $\mathbf{m} = [m_1, m_2, \dots, m_{MT}]$ is assigned, where the first $t$ values are set to $1$ and the remaining values to $0$, indicating the merging of the first $t$ LoRA branches. Here, $MT$ denotes the theoretically unlimited maximum number of tasks supported. The forward propagation for the $t$-th task is calculated as follows:
\begin{equation}
  \mathbf{e} = \mathbf{Wh} + \sum_{i=1}^t m_i \mathbf{\Delta W}_i \mathbf{h} = \mathbf{Wh} + \sum_{i=1}^t m_i \mathbf{A}_i \mathbf{B}_i \mathbf{h}.
  \label{eq:equation1}
\end{equation}

Note that during inference, the mask can be manually specified to adapt to different tasks. This enables the model to switch between tasks using the mask, eliminating the need to load multiple independent models.

\paragraph{\textbf{Theoretical Analysis}}
In continual learning, two critical characteristics are stability and plasticity. Stability focuses on retaining old knowledge throughout the learning process, while plasticity emphasizes the ability to learn new tasks. Achieving a trade-off between these two aspects is essential for successful continual learning. Our proposed MSLoRA module learns task-specific knowledge within a low-rank incremental subspace, thereby ensuring stability . Additionally, each task's learning relies on the original linear layer weights $\mathbf{W}$ rather than parameters from previously learned tasks, which reduces interference between tasks and enhances the model's plasticity. We substantiate this in the following proposition and proof.

\textbf{Proposition 1.} \textit{When learning the $t$-th task with forward propagation represented by \cref{eq:equation1}, fine-tuning $\mathbf{\Delta W}_t$ is equivalent to fine-tuning the pretrained weight $W$ and is independent of $\mathbf{\Delta W}_1$, $\mathbf{\Delta W}_2$, \ldots, $\mathbf{\Delta W}_{t-1}$.}

\textbf{Proof.} Based on \cref{eq:equation1} and the rule of differentiation, we can conclude that:
\begin{equation}
  \frac{\partial \mathbf{e}}{\partial \mathbf{W}} = \mathbf{h}^T, \frac{\partial \mathbf{e}}{\partial \mathbf{\Delta W}_i} = \mathbf{h}^T.
  \label{eq:equation2}
\end{equation}
When tuning the pretrained weight $W$ to learn the $t$-th task, we can compute the gradient of $W$ based on \cref{eq:equation2} and the chain rule:
\begin{equation}
  \frac{\partial \mathcal{L}}{\partial \mathbf{W}} = \frac{\partial \mathcal{L}}{\partial \mathbf{e}} \frac{\partial \mathbf{e}}{\partial \mathbf{W}} =  \frac{\partial \mathcal{L}}{\partial \mathbf{e}} \mathbf{h}^T.
  \label{eq:equation3}
\end{equation}

Similarly, according to \cref{eq:equation2} and \cref{eq:equation3}, when tuning the expanded weight $\mathbf{\Delta W}_t$, we can obtain the gradient of $\mathbf{\Delta W}_t$ as follows:
\begin{equation}
  \frac{\partial \mathcal{L}}{\partial \mathbf{\Delta W}_t} = \frac{\partial \mathcal{L}}{\partial \mathbf{e}} \frac{\partial \mathbf{e}}{\partial \mathbf{\Delta W}_t} =  \frac{\partial \mathcal{L}}{\partial \mathbf{e}} \mathbf{h}^T = \frac{\partial \mathcal{L}}{\partial \mathbf{W}}.
  \label{eq:equation4}
\end{equation}
Proposition 1 holds. Proposition 1 indicates that updating the LoRA weights for the $t$-th task is equivalent to updating the shared pretrained model weights $\mathbf{W}$, rather than building on the weights of the previous $t-1$ tasks. This characteristic reduces interference among the parameters of different tasks, enhancing plasticity. On the other hand, the LoRA-based fine-tuning method searches for task-specific parameters within a limited subspace based on the shared pretrained weights, ensuring stability. Therefore, by employing this module, our method achieves a balance between plasticity and stability.

\subsection{Contrastive Regularization Loss}
A key distinction between the MBIIL problem and traditional continual learning lies in the nature of the task sequences. In the MBIIL setting, tasks may involve the same or different biomedical image modalities, whereas traditional tasks typically focus on a single modality, such as natural images. The modality correlations or differences between tasks in MBIIL motivate us to employ the Contrastive Regularization (CR) method to better capture and leverage the modality-specific information across tasks.

\paragraph{\textbf{Desired Characteristics}}
When learning a specific task, we expect the parameters learned for that task to have a high degree of similarity with the parameters of other tasks that share the same modality, thereby promoting parameter sharing across tasks within the same modality. Conversely, tasks involving different modalities should exhibit lower parameter similarity, reducing conflict between tasks with different modalities.

\paragraph{\textbf{CR Loss}}
Based on the analysis above, given the current task $T_t$ and the set of previously learned tasks $\mathcal{T}_{\text{learned}} := \{T_1, T_2, \dots, T_{t-1}\}$, we partition $\mathcal{T}_{\text{learned}}$ into two subsets: $\mathcal{T}_{s} := \{T_i \mid T_i \in \mathcal{T}_{\text{learned}},\ a(T_i) = a(T_t)\}$ and $\mathcal{T}_{d} := \{T_i \mid T_i \in \mathcal{T}_{\text{learned}},\ a(T_i) \neq a(T_t)\}$, where $a(T_i)$ indicates the medical image modality associated with task $T_i$.
To learn task $T_t$, we introduce a new branch in the MSLoRA module, denoted as $w(T_t) = [\mathbf{A}_t, \mathbf{B}_t]$, where $w(T_i)$ represents the LoRA parameters for task $T_i$, and $\mathbf{A}_t$ and $\mathbf{B}_t$ are the two low-rank matrices specifically for task $T_t$. At this stage, $\text{MSLoRA} := \{w(T_i) \mid 1 \leq i \leq t\}$, with the first $t-1$ branches frozen and only the current branch $w(T_t)$ remaining trainable.

To build the CR loss function, we first define the similarity score between two parameter matrices in the LoRA using the following formula:
\begin{equation}
    \text{sim}(\mathbf{P}, \mathbf{Q}) = e^{-\textit{dis}(\mathbf{P}, \mathbf{Q})},
    \label{eq: equation5}
\end{equation}
where $\textit{dis}(\mathbf{P}, \mathbf{Q})$ denotes the Manhattan distance between matrix $\mathbf{P}$ and $\mathbf{Q}$.

Next, we regulate parameter updates within the LoRA module during training by applying a convergence loss for pairs within the same modality and a divergence loss for pairs across different modalities, as follows:
\begin{equation}
    \mathcal{L}_{\text{converge}} = \sum_{i=1}^{|\mathcal{T}_{s}|} 1-\text{sim}(w(T_i), w(T_t)),\ T_i \in \mathcal{T}_{s},
    \label{eq: equation6}
\end{equation}
\begin{equation}
    \mathcal{L}_{\text{diverge}} = \sum_{i=1}^{|\mathcal{T}_{d}|} \text{sim}(w(T_i), w(T_t)),\ T_i \in \mathcal{T}_{d}.
    \label{eq: equation7}
\end{equation}

Finally, the two terms are combined to form our CR loss, as follows:
\begin{equation}
    \mathcal{L}_{\text{cr}} = \mathcal{L}_{\text{converge}} + \mathcal{L}_{\text{diverge}}.
    \label{eq: equation8}
\end{equation}

This loss is added to the overall loss function, allowing us to leverage the knowledge learned from similar modality tasks by encouraging the parameters of the current task to be similar to those of previously learned tasks while ensuring they differ from the parameters of tasks with different modalities. This approach effectively reduces parameter conflicts during each forward propagation step.

\paragraph{\textbf{Orthogonal Loss}}
Some works, such as \cite{adalora}, indicate that incorporating orthogonality constraints into LoRA can effectively eliminate redundant parameters. To ensure the orthogonality of $\mathbf{A}_i$ and $\mathbf{B}_i$ in LoRA, we impose the conditions $\mathbf{A}_i^T \mathbf{A}_i = \mathbf{B}_i \mathbf{B}_i^T = \mathbf{I}$, where $\mathbf{I}$ is the identity matrix. To achieve this, we add the following regularization term during training:
\begin{equation}
    \mathcal{L}_{\text{ortho}} = \|\mathbf{A}_i^T \mathbf{A}_i - \mathbf{I}\|_F^2 +  \|\mathbf{B}_i^T \mathbf{B}_i - \mathbf{I}\|_F^2,
    \label{eq: equation9}
\end{equation}
where $\|\cdot\|_F^2$ denotes the squared Frobenius norm.

\subsection{The Complete Process of the MSLoRA-CR}
\cref{alg:alg1} outlines the complete process of our method. When the $t$-th task arrives, we expand a new branch in the MSLoRA module specifically for this task, fine-tuning this new branch while freezing the other branches. We augment the original loss function of the large vision language model with the CR loss and the orthogonal loss to help the model utilize the knowledge of previously learned tasks while learning the new task. Ultimately, our loss function is summarized as:
\begin{equation}
    \mathcal{L} = \mathcal{L}_{\text{ce}} + \alpha \mathcal{L}_{\text{cr}} + \beta \mathcal{L}_{\text{ortho}},
    \label{eq: equation10}
\end{equation}
where $\mathcal{L}_{\text{ce}}$ denotes the original loss of the large vision language model, $\alpha$ and $\beta$ are two hyper parameters.

\begin{algorithm}[t]
    \begin{algorithmic}[1]
        \STATE \textbf{Input:} The data of different modality-specific tasks $\{\mathcal{D}_i\}_{i=1}^T$. A pretrained LVLM $\mathcal{F}(\cdot)$.
        \STATE \textbf{Output:} MSLoRA modules with learned parameters $\{\mathbf{\Delta W}_i\}_{i=1}^T$.
        
        \STATE Initialize an empty MSLoRA modules set $\mathcal{M} = \{\}$;
        \FOR{$t = 1$ \textbf{ to } $T$}
            \STATE Expand a new LoRA branch $\mathbf{\Delta W}_t$ for the $t$-th task;
            \STATE $\mathcal{M} \gets \mathcal{M} \cup \{\mathbf{\Delta W}_t\}$;
            \FOR{each mini-batch $B_t$ sampled from $\mathcal{D}_t$}
                \STATE Compute the loss through \cref{eq: equation10};
                \STATE Perform backward propagation to update the parameters $\mathbf{\Delta W}_t$;
            \ENDFOR
        \ENDFOR
        
        \RETURN $\mathcal{M} = \{\mathbf{\Delta W}_i\}_{i=1}^T$ \textbf{ (learned parameters)};
    \end{algorithmic}
    \caption{ MSLoRA-CR for MBIIL}
    \label{alg:alg1}
\end{algorithm}

\section{Experiments}
\label{sec:experiments}
\subsection{Experimental Setting}

\begin{table}[ht]
  \centering
    \begin{tabular}{lcc}
    \specialrule{0.5pt}{0pt}{0pt}
    \toprule
    \multicolumn{1}{c}{\textbf{Datasets}} & \textbf{Modality} & \textbf{Task} \\
    \midrule
    PathVQA \cite{pathvqa} & Pathology & \multirow{4}[2]{*}{VQA} \\
    Slake \cite{slake} & Radiology &  \\
    VQA-Rad \cite{vqarad} & Radiology &  \\
    Fitzpatrick \cite{fitzpatrick} & Dermatology &  \\
    \midrule
    CXP \cite{cxp} & Radiology & \multirow{3}[2]{*}{Classification} \\
    PCam \cite{pcam} & Pathology &  \\
    HAM \cite{HAM} & Dermatology &  \\
    \midrule
    WSI-DX \cite{wsi-diagnosis} & Pathology & \multirow{2}[2]{*}{Report Generation} \\
    IU-X-Ray \cite{iu-x-ray} & Radiology &  \\
    \specialrule{0.5pt}{0pt}{0pt}
    \bottomrule
    \end{tabular}%
    
  \vspace{1mm}
  \caption{Overview of Dataset Tasks and Modalities.}
  \vspace{-4mm}
  \label{tab:dataset}
\end{table}%

\begin{table*}[ht]
  \centering
  \resizebox{\textwidth}{!}{
    \begin{tabular}{l|cc|cc|cc|cc|cc|c|cc|cc|c}
    \specialrule{0.5pt}{0pt}{0pt}
    \toprule
    \multicolumn{1}{c|}{\multirow{2}[1]{*}{\textbf{Models}}} & \multicolumn{2}{c|}{\textbf{PathVQA}} & \multicolumn{2}{c|}{\textbf{Slake-VQARad}} & \multicolumn{2}{c|}{\textbf{Fitzpatrick}} & \multicolumn{2}{c|}{\textbf{PCam}} & \multicolumn{2}{c|}{\textbf{CXP}} & \textbf{HAM} & \multicolumn{2}{c|}{\textbf{WSI-DX}} & \multicolumn{2}{c|}{\textbf{IU-X-Ray}} & \multirow{2}[1]{*}{\textbf{SUM}} \\
      & Open & Closed & Open & Closed & Open & Closed & F1-score & AUC & F1-score & AUC & ACC & F1-score & BLEU & F1-score & BLEU &  \\
    \midrule
    \multicolumn{16}{c}{\textbf{Previous Methods}} \\
    \midrule
    PathAsst(w/ CLIP)\cite{pathasst} & 37.60  & 89.70  & -  & -  & -  & -  & -  & -  & -  & -  & -  & -  & -  & -  & -  & -  \\
    Quilt-LLAVA\cite{quilt-llava} & 15.06  & 58.68  & 4.45  & 59.59  & 0.44  & 46.3  & 66.65  & 50.00  & 0  & 0.5  & 67.90  & 30.55  & 0.03  & 20.24  & 0.01  & 404.38  \\
     BiomedGPT-B\cite{BiomedGPT} & 28.00  & 88.00  & 48.01 & 36.34  & 0.12  & 10.01  & 0.79  & 49.85  & 43.22  & 51.84  & 3.30  & 14.26  & 0.01  & 15.82  & 0.05  & 345.88  \\
    
    
    LLaVA-Med-v1.5\cite{llava-med} $\dagger$ & 10.80  & 68.53  & 40.17  & 63.08  & 6.67  & 70.64  & 66.63  & 49.98  & 0.00  & 49.79  & 40.04  & 32.08  & 0.00  & 26.38  & 0.02  & 524.81  \\
    Data-Specific FT $\top$  & 37.44  & 91.54  & 73.17  & 79.36  & 56.98  & 92.25  & 91.83  & 91.90  & 90.89  & 85.34  & 77.68  & 47.99  & 9.41  & 40.62  & 3.89  & 970.29  \\
    \midrule
    \multicolumn{16}{c}{\textbf{MBIIL Methods}} \\
    \midrule
    \rowcolor{gray!20} MSLoRA & 37.44  & 91.54  & 72.59  & 80.38  & 54.73  & 91.94  & 92.74  & 92.80  & 88.05  & 79.96  & \textbf{82.03} & \textbf{52.87} & 14.20  & 40.97  & 4.67  & 976.89  \\
    \rowcolor{gray!20}MSLoRA-CR & 37.44  & 91.54  & \textbf{73.63} & \textbf{80.52} & \textbf{57.21} & \textbf{92.76} & 93.00  & 93.04  & 90.91  & 85.90  & 79.18  & 52.82  & 13.58  & 42.88  & 5.83  & 990.25 \\
    \rowcolor{gray!20}MSLoRA-CR-ORTHO & \textbf{37.44} & \textbf{91.54} & 73.40  & 80.09  & 50.96  & 89.73  & \textbf{93.21} & \textbf{93.26} & \textbf{91.05} & \textbf{87.31} & 78.23  & 52.67  & \textbf{14.74} & \textbf{47.47} & \textbf{14.12} & \textbf{995.21} \\
    \specialrule{0.5pt}{0pt}{0pt}
    \bottomrule
    \end{tabular}
  }
  \caption{
  Results (\%) of different methods per dataset and overall. $^\dagger$ marks the baseline, $\top$ the SOTA method. All results follow our unified evaluation except PathAsst, which is reported from the original paper due to lack of open-source code.
  }
  \vspace{-6mm}
  \label{tab:main}
\end{table*}

In this subsection, we provide the information of the datasets, evaluation metrics and implementation details.

\paragraph{\textbf{Datasets}}
To evaluate the MSLoRA-CR approach for the MBIIL task, we use nine datasets covering three modalities—pathology, radiology, and dermatology—and three tasks: Visual Question Answering (VQA), classification, and Report Generation, as detailed in \cref{tab:dataset}. Given that both Slake and VQA-Rad include radiology images for the VQA task, we merged them into a single dataset. We adapted the Fitzpatrick dataset from classification to VQA by leveraging its extensive labels. For classification, we reformulated each task into an instruction-based question-answering format, prompting the model to provide an open-ended response indicating the relevant category for each image. Further details on the datasets are provided in the supplementary materials.

\paragraph{\textbf{Evaluation Metrics}}
For VQA tasks, we evaluate model performance on closed-ended questions using accuracy and on open-ended questions using recall, as in previous studies \cite{llava, llava-med}. In the open-ended set, the model generates free-form text answers, while in the closed-ended set, responses are limited to predefined options.

For classification, we report the F1-score and AUC \cite{AUC}, except for the HAM seven-class task, where we report only accuracy. For report generation, we report both the F1-score and BLEU \cite{bleu} scores.
To assess overall performance on the MBIIL task across all tasks, we sum all task-specific metrics.

\paragraph{\textbf{Implementation Details}}
 We adopt LLaVA-Med-v1.5 \cite{llava-med} as the foundation model. All experiments are conducted on 8 A6000 GPUs (48GB each), with training and evaluation performed separately. We use the AdamW \cite{adamw} optimizer (no weight decay), a cosine learning rate scheduler, and a warm-up ratio of 0.03. Each experiment takes roughly 20 GPU hours, totaling around 500 GPU hours. Unless otherwise noted, both \textit{rank} and \textit{alpha} are set to 64.

\vspace{-0.5mm}
\subsection{Main Results}
As shown in \cref{tab:main}, Previous methods were often designed for specific medical imaging tasks or jointly fine-tuned on multiple modalities. However, they still lack generalization to unseen tasks and modalities. 

LLaVA-Med-v1.5, our baseline, was trained on extensive medical vision-language instruction tuning data (e.g., radiology and pathology), yet its zero-shot performance remains weak across tasks, indicating that simply increasing the diversity of training data does not guarantee strong generalization.
LoRA-Each represents the performance of separately fine-tuning a LoRA adapter for each modality, serving as the SOTA method for comparison. Theoretically, without considering modality relationships, this approach should yield the best possible results for individual tasks. However, MSLoRA demonstrates that even without additional regularization, dynamically expanding new branches for incoming tasks and merging them leads to superior performance compared to separate fine-tuning. This is likely due to an MoE-like (Mixture of Experts) effect, where merging multiple LoRA adapters facilitates knowledge sharing across modalities.
Furthermore, incorporating contrastive regularization (CR) boosts overall performance by approximately 16 points, verifying its effectiveness in enhancing task adaptation. The ORTHO loss further improves performance by reducing redundancy in LoRA parameters, which in turn reinforces contrastive regularization and enhances the model’s ability to generalize across biomedical imaging tasks. Notably, our model's performance is inherently constrained by the foundation model, as the primary source of biomedical knowledge comes from it. As a result, our approach still struggles to surpass specifically trained models on certain tasks. However, it still achieves competitive performance.

\begin{table*}[ht]
  \centering
  \resizebox{\textwidth}{!}{
    \begin{tabular}{l|cc|cc|cc|cc|cc|cc|c}
    \specialrule{0.5pt}{0pt}{0pt}
    \toprule
    \multicolumn{1}{c|}{\multirow{2}[2]{*}{\textbf{Models}}} & \multicolumn{2}{c|}{\textbf{PathVQA}} & \multicolumn{2}{c|}{\textbf{Slake-VQARad}} & \multicolumn{2}{c|}{\textbf{PCam}} & \multicolumn{2}{c|}{\textbf{CXP}} & \multicolumn{2}{c|}{\textbf{WSI-DX}} & \multicolumn{2}{c|}{\textbf{IU-X-Ray}} & \multirow{2}[2]{*}{\textbf{SUM}} \\
      & Open & Closed & Open & Closed & F1-score & AUC & F1-score & AUC & F1-score & BLEU & F1-score & BLEU &  \\
    \midrule
    MSLoRA-CR-ORTHO-8-8 & 38.18  & 92.72  & 73.42  & 80.23  & 91.51  & 91.59  & 91.04  & 86.03  & 51.67  & 13.16  & 38.86  & 4.83  & 753.24  \\
    MSLoRA-CR-ORTHO-16-16 & 38.09  & \textbf{93.31} & 72.51  & 81.25  & 92.10  & 92.19  & 90.60  & 85.17  & 51.48  & 13.46  & 43.24  & 10.23  & 763.61  \\
    \rowcolor{gray!20} MSLoRA-CR-ORTHO-32-32 & \textbf{38.29} & 92.36  & \textbf{73.67} & \textbf{81.40} & \textbf{92.96} & \textbf{93.01} & 90.77  & 85.77  & 52.47  & 13.75  & \textbf{45.99} & \textbf{13.09} & \textbf{773.54} \\
    MSLoRA-CR-ORTHO-64-64 & 37.44  & 91.54  & 72.62  & 80.96  & 92.66  & 92.71  & \textbf{91.75} & \textbf{87.86} & \textbf{53.11} & 13.74  & 41.41  & 5.04  & 760.83  \\
    MSLoRA-CR-ORTHO-128-128 & 27.40  & 74.58  & 63.65  & 61.77  & 92.71  & 92.77  & 84.26  & 74.49  & 52.52  & \textbf{14.62} & 36.16  & 4.01  & 678.94  \\
    \bottomrule
    \specialrule{0.5pt}{0pt}{0pt}
    \end{tabular}
 }
  \caption{Results (\%) under varying LoRA hyperparameters. The two numbers in the model name indicate the LoRA hyperparameters, specifically rank and alpha, respectively.}
  
  \vspace{-4mm}
  \label{tab:lora_rank}
\end{table*}

  

\begin{table*}[ht]
  \centering
  \resizebox{\textwidth}{!}{
    \begin{tabular}{c|c|c|cc|cc|cc|cc|cc|cc|c}
    \specialrule{0.5pt}{0pt}{0pt}
    \toprule
    \multirow{2}[2]{*}{$\alpha$} & \multirow{2}[2]{*}{$\beta$} & \multirow{2}[2]{*}{\textbf{trainable}} & \multicolumn{2}{c|}{\textbf{PathVQA}} & \multicolumn{2}{c|}{\textbf{Slake-VQARad}} & \multicolumn{2}{c|}{\textbf{PCam}} & \multicolumn{2}{c|}{\textbf{CXP}} & \multicolumn{2}{c|}{\textbf{WSI-DX}} & \multicolumn{2}{c|}{\textbf{IU-X-Ray}} & \multirow{2}[2]{*}{\textbf{SUM}} \\
      &   &   & Open & Closed & Open & Closed & F1-score & AUC & F1-score & AUC & F1-score & BLEU & F1-score & BLEU &  \\
    \midrule
    0.1 & 0.1 & TRUE & 37.44  & 91.54  & 72.36  & 80.67  & \textbf{93.07} & \textbf{93.10} & 87.43  & 78.68  & 52.50  & \textbf{14.17} & 39.85  & 3.85  & 744.65  \\
    0.01 & 0.01 & TRUE & 37.44  & 91.54  & 73.21  & 80.67  & 93.06  & 93.10  & 90.09  & 84.57  & 52.66  & 14.16  & 41.00  & 4.96  & 756.46  \\
    0.01 & 0.01 & FALSE & 37.44  & 91.54  & \textbf{73.51} & 80.23  & 92.69  & 92.74  & 89.40  & 83.38  & 52.19  & 14.16  & \textbf{41.93} & \textbf{5.42} & 754.63  \\
    0.1 & 0.1 & FALSE & 37.44  & 91.54  & 71.86  & 79.94  & 93.05  & 93.10  & 88.51  & 81.75  & 52.16  & 13.75  & 41.84  & 5.14  & 750.08  \\
    0.01 & 0.1 & FALSE & 37.44  & 91.54  & 73.40  & 80.09  & 92.94  & 93.00  & 90.82  & 85.60  & \textbf{53.16} & 14.14  & 40.88  & 4.36  & 757.36  \\
    \rowcolor{gray!20} 0.1 & 0.01 & FALSE & \textbf{37.44} & \textbf{91.54} & 72.62  & \textbf{80.96} & 92.66  & 92.71  & \textbf{91.75} & \textbf{87.86} & 53.11  & 13.74  & 41.41  & 5.04  & \textbf{760.83} \\
    \bottomrule
    \specialrule{0.5pt}{0pt}{0pt}
    \end{tabular}
  }
  \caption{Detailed results (\%) for varying $\alpha$ and $\beta$.}
  \vspace{-4mm}
  \label{tab:alpha_beta}
\end{table*}

\begin{table*}[ht]
  \centering
  \resizebox{\textwidth}{!}{
    \begin{tabular}{l|cc|cc|cc|cc|cc|cc|c}
    \specialrule{0.5pt}{0pt}{0pt}
    \toprule
    \multicolumn{1}{c|}{\multirow{2}[2]{*}{\textbf{Models}}} & \multicolumn{2}{c|}{\textbf{PathVQA}} & \multicolumn{2}{c|}{\textbf{Slake-VQARad}} & \multicolumn{2}{c|}{\textbf{PCam}} & \multicolumn{2}{c|}{\textbf{CXP}} & \multicolumn{2}{c|}{\textbf{WSI-DX}} & \multicolumn{2}{c|}{\textbf{IU-X-Ray}} & \multirow{2}[2]{*}{\textbf{SUM}} \\
      & Open & Closed & Open & Closed & F1-score & AUC & F1-score & AUC & F1-score & BLEU & F1-score & BLEU &  \\
    \midrule
    MSLoRA-CR-ORTHO-deep & 34.51  & 90.53  & 71.32  & 78.78  & 90.62  & 90.68  & 87.12  & 76.54  & 51.78  & 13.48  & 39.94  & 4.79  & 730.08  \\
    MSLoRA-CR-ORTHO-shallow & 35.68  & \textbf{91.86} & \textbf{73.19} & 79.94  & 91.39  & 91.46  & \textbf{92.63} & \textbf{90.56} & 50.42  & 12.58  & 38.58  & 4.41  & 752.70  \\
    \rowcolor{gray!20} MSLoRA-CR-ORTHO-all & \textbf{37.44} & 91.54  & 72.62  & \textbf{80.96} & \textbf{92.66} & \textbf{92.71} & 91.75  & 87.86  & \textbf{53.11} & \textbf{13.74} & \textbf{41.41} & \textbf{5.04} & \textbf{760.83} \\
    \bottomrule
    \specialrule{0.5pt}{0pt}{0pt}
    \end{tabular}
  }
  \caption{ Results (\%) illustrating the impact of the MSLoRA module placement on model performance across various tasks. The ``deep'' model adds our module to the deepest 8 layers of the language model, the ``shallow'' model integrates the module into the shallowest 8 layers, and the ``all'' model applies the module across all layers.}
  \vspace{-5mm}
  \label{tab:placement}
\end{table*}

\begin{table*}[ht]
  \centering
  \resizebox{\textwidth}{!}{
    \begin{tabular}{l|c|c|c|c|c|c|c|c|c|c|c|c|c}
    \specialrule{0.5pt}{0pt}{0pt}
    \toprule
    \multicolumn{1}{c|}{\multirow{2}[2]{*}{\textbf{Order}}} & \multicolumn{2}{c|}{\textbf{PathVQA}} & \multicolumn{2}{c|}{\textbf{Slake-VQARad}} & \multicolumn{2}{c|}{\textbf{PCam}} & \multicolumn{2}{c|}{\textbf{CXP}} & \multicolumn{2}{c|}{\textbf{WSI-DX}} & \multicolumn{2}{c|}{\textbf{IU-X-Ray}} & \multirow{2}[2]{*}{\textbf{SUM}} \\
          & \multicolumn{1}{c}{Open} & Closed & \multicolumn{1}{c}{Open} & Closed & \multicolumn{1}{c}{F1-score} & AUC   & \multicolumn{1}{c}{F1-score} & AUC   & \multicolumn{1}{c}{F1-score} & BLEU  & \multicolumn{1}{c}{F1-score} & BLEU  &  \\
    \midrule
    $A_1,A_2,A_3,B_1,B_2,B_3$ & \textbf{37.45 } & \textbf{91.54 } & 72.11  & 79.94  & 92.33  & 92.42  & 88.97  & 81.97  & 52.02  & 13.20  & 41.49  & 5.09  & 748.52  \\
    $B_1,B_2,B_3,A_1,A_2,A_3$ & 8.53  & 73.52  & 72.58  & 79.22  & 88.79  & 88.77  & 90.67  & 84.91  & 40.49  & 7.93  & \textbf{41.79 } & 5.05  & 682.25  \\
    $A_2,B_2,A_1,B_1,A_3,B_3$ & 5.46  & 62.02  & 58.44  & 56.10  & 92.05  & 92.08  & 90.35  & 85.47  & 48.51  & 12.39  & 35.73  & 3.69  & 642.30  \\
    $A_3,B_3,A_2,B_2,A_1,B_1$ & 5.57  & 74.70  & 17.03  & 46.80  & \textbf{93.48 } & \textbf{93.50 } & 86.91  & 77.14  & \textbf{53.63 } & \textbf{14.32 } & 41.65  & \textbf{5.69 } & 610.40  \\
    \rowcolor{gray!20} $A_1,B_1,A_2,B_2,A_3,B_3$ & 37.44  & \textbf{91.54 } & \textbf{72.62 } & \textbf{80.96 } & 92.66  & 92.71  & \textbf{91.75 } & \textbf{87.86 } & 53.11  & 13.74  & 41.41  & 5.04  & \textbf{760.83 } \\
    \bottomrule
    \specialrule{0.5pt}{0pt}{0pt}
    \end{tabular}
  }
  \caption{Comparison of performance (\%) across different modality-specific task sequence orders. $A_1, B_1, A_2, B_2, A_3, B_3$ correspond to PathVQA, Slake-VQARad, PCam, CXP, WSI-DX, and IU-X-Ray, respectively, where $A$ denotes pathology and $B$ denotes radiology, with subscripts $1, 2, 3$ indicating VQA, classification, and report generation tasks.}

  \label{tab:order}
  \vspace{-6mm}
\end{table*}

\begin{table*}[ht]
  \centering
  \resizebox{\textwidth}{!}{
    \begin{tabular}{cp{10.89em}p{36.335em}}
    \specialrule{0.5pt}{0pt}{0pt}
    \toprule
    \multicolumn{1}{c}{\multirow{4}[2]{*}{\includegraphics[width=2cm]{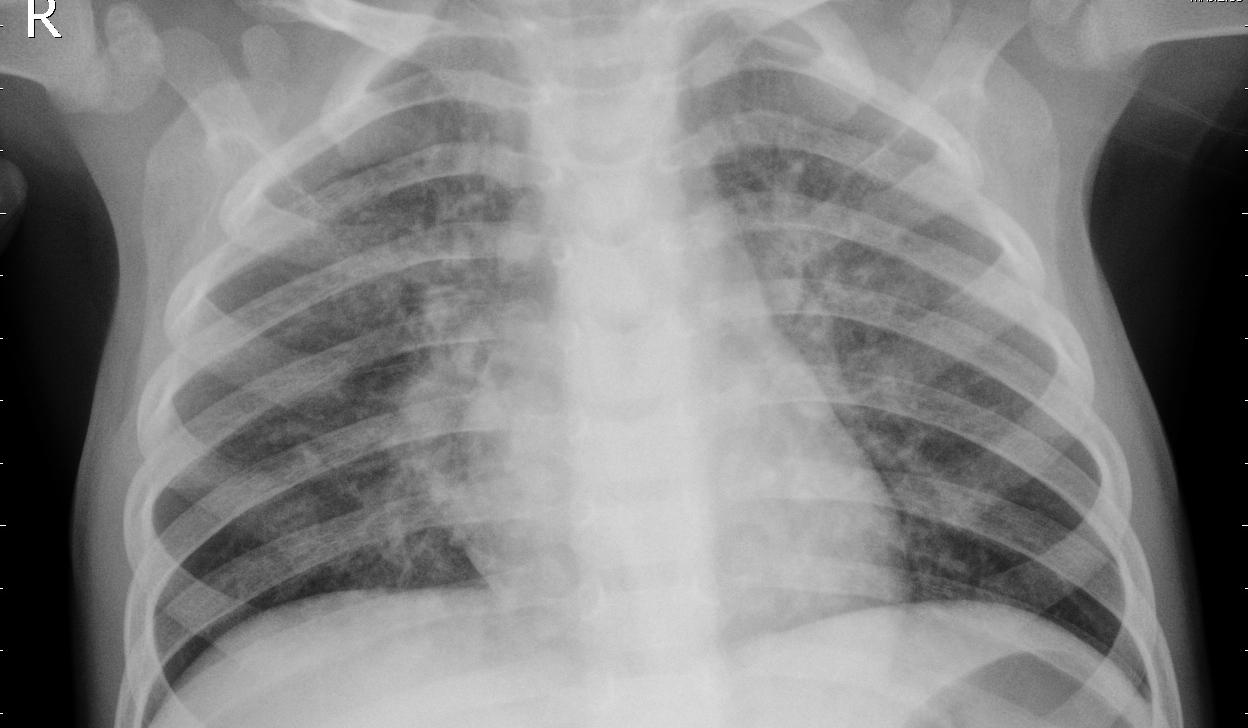}}} & User: & Given a chest X-ray image, determine whether the patient has a normal chest or pneumonia. Only output `normal' or `pneumonia'. \\
      & LLaVA-Med-v1.5: & \textbf{ONLY ANSWER:} The chest x-ray image shows a normal chest. \\
      & Ours: & Pneumonia. \\
      & Ground truth:  & Pneumonia. \\
    \midrule
    \multicolumn{1}{c}{\multirow{4}[2]{*}{\includegraphics[width=2cm]{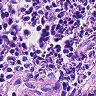}}} & User: & Given a color image extracted from histopathologic scans of lymph node sections,
    analyze the image for the presence of metastatic tissue. Classify the image into ... \\
      & LLaVA-Med-v1.5: &\textbf{ONLY ANSWER:} The image is classified as `positive' because it contains at least one pixel of tumor tissue in the center 32x32px region. \\
      & Ours: & Negative. \\
      & Ground truth:  & Negative. \\
    \bottomrule
    \specialrule{0.5pt}{0pt}{0pt}
    \end{tabular}
    }
  \caption{Comparison of Our Method with LLaVA-Med-v1.5 on Classification Task. ``\textbf{ONLY ANSWER:}'' indicates that the model responds exclusively with the following content across all test samples.}
  \vspace{-5mm}
  \label{tab:example}
\end{table*}

\begin{figure*}[ht]
    \centering
    \includegraphics[width=1\linewidth]{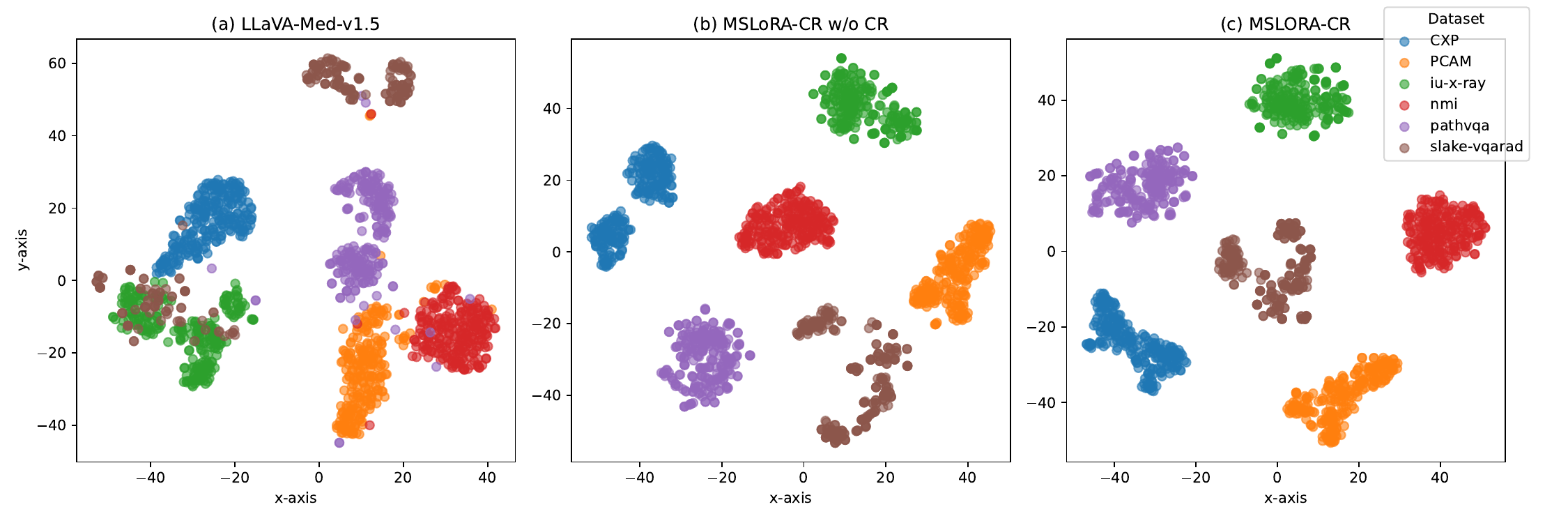}
    \vspace{-8mm}
    \caption{t-SNE visualization of the vision embeddings from three different models. Each subplot corresponds to a different model: (a) LLaVA-Med-v1.5, (b) MSLoRA-CR w/o CR, and (c) MSLoRA-CR.}
    \vspace{-4mm}
    \label{fig:tsne}
\end{figure*}

\vspace{-0.5mm}
\subsection{Ablation Study}
We perform ablation studies on 6 representative datasets—PathVQA, Slake-VQARad, PCam, CXP, WSI-DX, and IU-X-Ray—to ensure computational efficiency and modality diversity.

\paragraph{\textbf{Effect of LoRA rank}}
As shown in \cref{tab:lora_rank}, it is evident that the model performance improves as the LoRA rank increases initially. However, after reaching a certain point, further increases in rank lead to a decline in performance. This drop may be attributed to the increased model complexity, which could slow down the convergence of the orthogonal loss and contrastive regularization loss, ultimately leading to suboptimal results.

\paragraph{\textbf{Effect of hyper-parameters $\alpha$ and $\beta$}}
Exhaustive grid search over all $\alpha$ and $\beta$ combinations is impractical due to high computational costs, so we selected a few combinations based on heuristic principles to assess the impact on model performance. Our empirical search process was as follows: we first trained for several steps to observe the scale differences among the loss terms, then adjusted $\alpha$ and $\beta$ to ensure that all three loss components were on a similar scale. We found that when $\alpha = 0.1$ and $\beta = 0.01$, the loss terms were balanced, leading to the best performance. As shown in \cref{tab:alpha_beta}, deviations from this balance resulted in suboptimal outcomes, and allowing $\alpha$ and $\beta$ to be trainable made it difficult to find the optimal values. Therefore, we adopted this fixed setting. A more detailed analysis is provided in the supplementary materials.

\paragraph{\textbf{Effect of MSLoRA module placement}} 
\cref{tab:placement} suggests that adding the MSLoRA module only to the shallowest 8 layers yields higher classification performance on the CXP task, likely because introducing fewer parameters early in encoding can mitigate overfitting. However, both shallow and deep configurations underperform in the report generation task compared to adding modules across all layers. This disparity may arise because limiting the module to specific layers restricts the language model's capacity to thoroughly capture and represent intricate, long-form sentence structures, which is essential for generating coherent and accurate reports.

\paragraph{\textbf{Effect of Task Sequence order}} 
 \cref{tab:order} compares the performance (\%) of different modality-specific task sequence orders. In this context, $A_1$, $B_1$, $A_2$, $B_2$, $A_3$, and $B_3$ correspond to PathVQA, Slake-VQARad, PCam, CXP, WSI-DX, and IU-X-Ray, respectively. Here, $A$ represents the pathology modality, $B$ represents the radiology modality, while $1$, $2$, and $3$ indicate the VQA, classification, and report generation tasks, respectively.
 
From the first two rows, it is evident that the order in which modality tasks are trained significantly affects model performance. Specifically, training the pathology modality tasks first, followed by the radiology modality tasks, results in better performance than when the training sequence is reversed. This discrepancy is likely due to the larger size of the pathology modality dataset, which contains 48,332 question-answer pairs, compared to the 14,597 pairs in the radiology modality dataset.

The last three rows of the table highlight the effect of different task training sequences on performance. Notably, training the report generation task first leads to the worst performance. When the classification task is trained first, performance on both VQA and report generation tasks declines. Conversely, training the VQA task first results in the best overall performance. This phenomenon suggests that the task trained first has a significant impact on our model performance. The VQA task, with its diverse set of possible answers, helps the model learn more generalizable patterns and better adhere to human instructions. In contrast, the report generation and classification tasks, which require the model to generate fixed templates or restricted phrases, limit the model's ability to generalize and follow instructions effectively.

\subsection{Qualitative Analysis}
\paragraph{\textbf{Case study:}}
\cref{tab:example} presents a case on the classification task. While LLaVA-Med-v1.5, trained on multiple modalities, often produces identical responses and struggles to generalize effectively to specific downstream tasks, our approach demonstrates better adaptability. Additional examples across various tasks and modalities, including failure cases, are available in the supplementary materials.
\paragraph{\textbf{Visualization Analysis}}
\cref{fig:tsne} shows t-SNE visualizations of the visual embeddings from three methods. We extracted the visual token embeddings from the last hidden layer and averaged them, using a balanced sample of 250 examples per dataset. These embeddings reflect how the language model utilizes visual information to handle specific tasks. LLaVA-Med V1.5 exhibits limited ability to distinguish between tasks, while MSLoRA-CR w/o CR achieves better separation. Additionally, the use of contrastive loss improves intra-class clustering, as seen in the CXP dataset, which corresponds to the performance gains in \cref{tab:main}.

\section{Conclusion}
In this paper, we propose the modality-specific LoRA fine-tuning with contrastive regularization (MSLoRA-CR) approach to tackle the multimodal biomedical image incremental learning (MBIIL) problem. We introduce and theoretically analyze new LoRA branches within the MSLoRA module to enable modality-specific task learning, demonstrating that this method achieves an optimal balance between stability and plasticity.
The CR loss further strengthens learning by emphasizing associations or distinctions between modalities. Experimental results across multiple datasets and tasks validate the effectiveness of our approach. Future research will explore integrating data incremental learning into this framework, as well as extending it to support detection and segmentation tasks. Moreover, natural scenes exhibit domain differences, such as real-world images, artwork, and illustrations. While existing large vision-language models focus on unified joint training, our method demonstrates stronger applicability and ease of use in natural scenes.

\clearpage
\begin{acks}
This work was supported by the National Key R\&D Program of China (2024YFB4105200), the National Natural Science Foundation of China (Grant No. 62473384, 62473385, U24A20256, and 62202499), the Beijing Municipal Natural Science Foundation (Grant Number: L231012). We are grateful to the High Performance Computing Center of Central South University for partial support of this work.
\end{acks}

\bibliographystyle{ACM-Reference-Format}
\balance
\bibliography{main}


\begin{thebibliography}{58}


\ifx \showCODEN    \undefined \def \showCODEN     #1{\unskip}     \fi
\ifx \showDOI      \undefined \def \showDOI       #1{#1}\fi
\ifx \showISBNx    \undefined \def \showISBNx     #1{\unskip}     \fi
\ifx \showISBNxiii \undefined \def \showISBNxiii  #1{\unskip}     \fi
\ifx \showISSN     \undefined \def \showISSN      #1{\unskip}     \fi
\ifx \showLCCN     \undefined \def \showLCCN      #1{\unskip}     \fi
\ifx \shownote     \undefined \def \shownote      #1{#1}          \fi
\ifx \showarticletitle \undefined \def \showarticletitle #1{#1}   \fi
\ifx \showURL      \undefined \def \showURL       {\relax}        \fi
\providecommand\bibfield[2]{#2}
\providecommand\bibinfo[2]{#2}
\providecommand\natexlab[1]{#1}
\providecommand\showeprint[2][]{arXiv:#2}

\bibitem[Alayrac et~al\mbox{.}(2022)]%
        {flamingo}
\bibfield{author}{\bibinfo{person}{Jean-Baptiste Alayrac}, \bibinfo{person}{Jeff Donahue}, \bibinfo{person}{Pauline Luc}, \bibinfo{person}{Antoine Miech}, \bibinfo{person}{Iain Barr}, \bibinfo{person}{Yana Hasson}, \bibinfo{person}{Karel Lenc}, \bibinfo{person}{Arthur Mensch}, \bibinfo{person}{Katherine Millican}, \bibinfo{person}{Malcolm Reynolds}, {et~al\mbox{.}}} \bibinfo{year}{2022}\natexlab{}.
\newblock \showarticletitle{Flamingo: a visual language model for few-shot learning}.
\newblock \bibinfo{journal}{\emph{Advances in neural information processing systems}}  \bibinfo{volume}{35} (\bibinfo{year}{2022}), \bibinfo{pages}{23716--23736}.
\newblock


\bibitem[Aljundi et~al\mbox{.}(2018)]%
        {cl-r-mas}
\bibfield{author}{\bibinfo{person}{Rahaf Aljundi}, \bibinfo{person}{Francesca Babiloni}, \bibinfo{person}{Mohamed Elhoseiny}, \bibinfo{person}{Marcus Rohrbach}, {and} \bibinfo{person}{Tinne Tuytelaars}.} \bibinfo{year}{2018}\natexlab{}.
\newblock \showarticletitle{Memory aware synapses: Learning what (not) to forget}. In \bibinfo{booktitle}{\emph{Proceedings of the European conference on computer vision (ECCV)}}. \bibinfo{pages}{139--154}.
\newblock


\bibitem[Aljundi et~al\mbox{.}(2017)]%
        {cl-a-expert-gate}
\bibfield{author}{\bibinfo{person}{Rahaf Aljundi}, \bibinfo{person}{Punarjay Chakravarty}, {and} \bibinfo{person}{Tinne Tuytelaars}.} \bibinfo{year}{2017}\natexlab{}.
\newblock \showarticletitle{Expert gate: Lifelong learning with a network of experts}. In \bibinfo{booktitle}{\emph{Proceedings of the IEEE conference on computer vision and pattern recognition}}. \bibinfo{pages}{3366--3375}.
\newblock


\bibitem[Douillard et~al\mbox{.}(2022)]%
        {cl-a-dytox}
\bibfield{author}{\bibinfo{person}{Arthur Douillard}, \bibinfo{person}{Alexandre Ram{\'e}}, \bibinfo{person}{Guillaume Couairon}, {and} \bibinfo{person}{Matthieu Cord}.} \bibinfo{year}{2022}\natexlab{}.
\newblock \showarticletitle{Dytox: Transformers for continual learning with dynamic token expansion}. In \bibinfo{booktitle}{\emph{Proceedings of the IEEE/CVF Conference on Computer Vision and Pattern Recognition}}. \bibinfo{pages}{9285--9295}.
\newblock


\bibitem[Gao et~al\mbox{.}(2024)]%
        {clip-adapter}
\bibfield{author}{\bibinfo{person}{Peng Gao}, \bibinfo{person}{Shijie Geng}, \bibinfo{person}{Renrui Zhang}, \bibinfo{person}{Teli Ma}, \bibinfo{person}{Rongyao Fang}, \bibinfo{person}{Yongfeng Zhang}, \bibinfo{person}{Hongsheng Li}, {and} \bibinfo{person}{Yu Qiao}.} \bibinfo{year}{2024}\natexlab{}.
\newblock \showarticletitle{Clip-adapter: Better vision-language models with feature adapters}.
\newblock \bibinfo{journal}{\emph{International Journal of Computer Vision}} \bibinfo{volume}{132}, \bibinfo{number}{2} (\bibinfo{year}{2024}), \bibinfo{pages}{581--595}.
\newblock


\bibitem[Groh et~al\mbox{.}(2021)]%
        {fitzpatrick}
\bibfield{author}{\bibinfo{person}{Matthew Groh}, \bibinfo{person}{Caleb Harris}, \bibinfo{person}{Luis Soenksen}, \bibinfo{person}{Felix Lau}, \bibinfo{person}{Rachel Han}, \bibinfo{person}{Aerin Kim}, \bibinfo{person}{Arash Koochek}, {and} \bibinfo{person}{Omar Badri}.} \bibinfo{year}{2021}\natexlab{}.
\newblock \showarticletitle{Evaluating deep neural networks trained on clinical images in dermatology with the fitzpatrick 17k dataset}. In \bibinfo{booktitle}{\emph{Proceedings of the IEEE/CVF Conference on Computer Vision and Pattern Recognition}}. \bibinfo{pages}{1820--1828}.
\newblock


\bibitem[Hanley and McNeil(1982)]%
        {AUC}
\bibfield{author}{\bibinfo{person}{James~A Hanley} {and} \bibinfo{person}{Barbara~J McNeil}.} \bibinfo{year}{1982}\natexlab{}.
\newblock \showarticletitle{The meaning and use of the area under a receiver operating characteristic (ROC) curve.}
\newblock \bibinfo{journal}{\emph{Radiology}} \bibinfo{volume}{143}, \bibinfo{number}{1} (\bibinfo{year}{1982}), \bibinfo{pages}{29--36}.
\newblock


\bibitem[He et~al\mbox{.}(2020)]%
        {pathvqa}
\bibfield{author}{\bibinfo{person}{Xuehai He}, \bibinfo{person}{Yichen Zhang}, \bibinfo{person}{Luntian Mou}, \bibinfo{person}{Eric Xing}, {and} \bibinfo{person}{Pengtao Xie}.} \bibinfo{year}{2020}\natexlab{}.
\newblock \showarticletitle{PathVQA: 30000+ Questions for Medical Visual Question Answering}.
\newblock \bibinfo{journal}{\emph{arXiv preprint arXiv:2003.10286}} (\bibinfo{year}{2020}).
\newblock


\bibitem[Houlsby et~al\mbox{.}(2019)]%
        {adapter}
\bibfield{author}{\bibinfo{person}{Neil Houlsby}, \bibinfo{person}{Andrei Giurgiu}, \bibinfo{person}{Stanislaw Jastrzebski}, \bibinfo{person}{Bruna Morrone}, \bibinfo{person}{Quentin De~Laroussilhe}, \bibinfo{person}{Andrea Gesmundo}, \bibinfo{person}{Mona Attariyan}, {and} \bibinfo{person}{Sylvain Gelly}.} \bibinfo{year}{2019}\natexlab{}.
\newblock \showarticletitle{Parameter-efficient transfer learning for NLP}. In \bibinfo{booktitle}{\emph{International conference on machine learning}}. PMLR, \bibinfo{pages}{2790--2799}.
\newblock


\bibitem[Hu et~al\mbox{.}(2021)]%
        {lora}
\bibfield{author}{\bibinfo{person}{Edward~J Hu}, \bibinfo{person}{Yelong Shen}, \bibinfo{person}{Phillip Wallis}, \bibinfo{person}{Zeyuan Allen-Zhu}, \bibinfo{person}{Yuanzhi Li}, \bibinfo{person}{Shean Wang}, \bibinfo{person}{Lu Wang}, {and} \bibinfo{person}{Weizhu Chen}.} \bibinfo{year}{2021}\natexlab{}.
\newblock \showarticletitle{Lora: Low-rank adaptation of large language models}.
\newblock \bibinfo{journal}{\emph{arXiv preprint arXiv:2106.09685}} (\bibinfo{year}{2021}).
\newblock


\bibitem[Hu et~al\mbox{.}(2023b)]%
        {cl-a-DNE}
\bibfield{author}{\bibinfo{person}{Zhiyuan Hu}, \bibinfo{person}{Yunsheng Li}, \bibinfo{person}{Jiancheng Lyu}, \bibinfo{person}{Dashan Gao}, {and} \bibinfo{person}{Nuno Vasconcelos}.} \bibinfo{year}{2023}\natexlab{b}.
\newblock \showarticletitle{Dense network expansion for class incremental learning}. In \bibinfo{booktitle}{\emph{Proceedings of the IEEE/CVF Conference on Computer Vision and Pattern Recognition}}. \bibinfo{pages}{11858--11867}.
\newblock


\bibitem[Hu et~al\mbox{.}(2023a)]%
        {intro-peft2}
\bibfield{author}{\bibinfo{person}{Zi-Yuan Hu}, \bibinfo{person}{Yanyang Li}, \bibinfo{person}{Michael~R Lyu}, {and} \bibinfo{person}{Liwei Wang}.} \bibinfo{year}{2023}\natexlab{a}.
\newblock \showarticletitle{Vl-pet: Vision-and-language parameter-efficient tuning via granularity control}. In \bibinfo{booktitle}{\emph{Proceedings of the IEEE/CVF International Conference on Computer Vision}}. \bibinfo{pages}{3010--3020}.
\newblock


\bibitem[Huang et~al\mbox{.}(2025)]%
        {hira}
\bibfield{author}{\bibinfo{person}{Qiushi Huang}, \bibinfo{person}{Tom Ko}, \bibinfo{person}{Zhan Zhuang}, \bibinfo{person}{Lilian Tang}, {and} \bibinfo{person}{Yu Zhang}.} \bibinfo{year}{2025}\natexlab{}.
\newblock \showarticletitle{HiRA: Parameter-Efficient Hadamard High-Rank Adaptation for Large Language Models}. In \bibinfo{booktitle}{\emph{The Thirteenth International Conference on Learning Representations}}.
\newblock


\bibitem[Isele and Cosgun(2018)]%
        {cl-m-replay}
\bibfield{author}{\bibinfo{person}{David Isele} {and} \bibinfo{person}{Akansel Cosgun}.} \bibinfo{year}{2018}\natexlab{}.
\newblock \showarticletitle{Selective experience replay for lifelong learning}. In \bibinfo{booktitle}{\emph{Proceedings of the AAAI Conference on Artificial Intelligence}}, Vol.~\bibinfo{volume}{32}.
\newblock


\bibitem[Jia et~al\mbox{.}(2022)]%
        {intro-peft}
\bibfield{author}{\bibinfo{person}{Menglin Jia}, \bibinfo{person}{Luming Tang}, \bibinfo{person}{Bor-Chun Chen}, \bibinfo{person}{Claire Cardie}, \bibinfo{person}{Serge Belongie}, \bibinfo{person}{Bharath Hariharan}, {and} \bibinfo{person}{Ser-Nam Lim}.} \bibinfo{year}{2022}\natexlab{}.
\newblock \showarticletitle{Visual prompt tuning}. In \bibinfo{booktitle}{\emph{European Conference on Computer Vision}}. Springer, \bibinfo{pages}{709--727}.
\newblock


\bibitem[Jiang et~al\mbox{.}(2025)]%
        {jiang2025unlocking}
\bibfield{author}{\bibinfo{person}{Gangwei Jiang}, \bibinfo{person}{Caigao Jiang}, \bibinfo{person}{Zhaoyi Li}, \bibinfo{person}{Siqiao Xue}, \bibinfo{person}{Jun Zhou}, \bibinfo{person}{Linqi Song}, \bibinfo{person}{Defu Lian}, {and} \bibinfo{person}{Yin Wei}.} \bibinfo{year}{2025}\natexlab{}.
\newblock \showarticletitle{Unlocking the Power of Function Vectors for Characterizing and Mitigating Catastrophic Forgetting in Continual Instruction Tuning}.
\newblock \bibinfo{journal}{\emph{arXiv preprint arXiv:2502.11019}} (\bibinfo{year}{2025}).
\newblock


\bibitem[Kermany et~al\mbox{.}(2018)]%
        {cxp}
\bibfield{author}{\bibinfo{person}{Daniel~S Kermany}, \bibinfo{person}{Michael Goldbaum}, \bibinfo{person}{Wenjia Cai}, \bibinfo{person}{Carolina~CS Valentim}, \bibinfo{person}{Huiying Liang}, \bibinfo{person}{Sally~L Baxter}, \bibinfo{person}{Alex McKeown}, \bibinfo{person}{Ge Yang}, \bibinfo{person}{Xiaokang Wu}, \bibinfo{person}{Fangbing Yan}, {et~al\mbox{.}}} \bibinfo{year}{2018}\natexlab{}.
\newblock \showarticletitle{Identifying medical diagnoses and treatable diseases by image-based deep learning}.
\newblock \bibinfo{journal}{\emph{cell}} \bibinfo{volume}{172}, \bibinfo{number}{5} (\bibinfo{year}{2018}), \bibinfo{pages}{1122--1131}.
\newblock


\bibitem[Kim et~al\mbox{.}(2024)]%
        {hydra-lora}
\bibfield{author}{\bibinfo{person}{Sanghyeon Kim}, \bibinfo{person}{Hyunmo Yang}, \bibinfo{person}{Yunghyun Kim}, \bibinfo{person}{Youngjoon Hong}, {and} \bibinfo{person}{Eunbyung Park}.} \bibinfo{year}{2024}\natexlab{}.
\newblock \showarticletitle{Hydra: Multi-head low-rank adaptation for parameter efficient fine-tuning}.
\newblock \bibinfo{journal}{\emph{Neural Networks}}  \bibinfo{volume}{178} (\bibinfo{year}{2024}), \bibinfo{pages}{106414}.
\newblock


\bibitem[Kirkpatrick et~al\mbox{.}(2017)]%
        {cl-r-ewc}
\bibfield{author}{\bibinfo{person}{James Kirkpatrick}, \bibinfo{person}{Razvan Pascanu}, \bibinfo{person}{Neil Rabinowitz}, \bibinfo{person}{Joel Veness}, \bibinfo{person}{Guillaume Desjardins}, \bibinfo{person}{Andrei~A Rusu}, \bibinfo{person}{Kieran Milan}, \bibinfo{person}{John Quan}, \bibinfo{person}{Tiago Ramalho}, \bibinfo{person}{Agnieszka Grabska-Barwinska}, {et~al\mbox{.}}} \bibinfo{year}{2017}\natexlab{}.
\newblock \showarticletitle{Overcoming catastrophic forgetting in neural networks}.
\newblock \bibinfo{journal}{\emph{Proceedings of the national academy of sciences}} \bibinfo{volume}{114}, \bibinfo{number}{13} (\bibinfo{year}{2017}), \bibinfo{pages}{3521--3526}.
\newblock


\bibitem[Lau et~al\mbox{.}(2018)]%
        {vqarad}
\bibfield{author}{\bibinfo{person}{Jason~J Lau}, \bibinfo{person}{Soumya Gayen}, \bibinfo{person}{Asma Ben~Abacha}, {and} \bibinfo{person}{Dina Demner-Fushman}.} \bibinfo{year}{2018}\natexlab{}.
\newblock \showarticletitle{A dataset of clinically generated visual questions and answers about radiology images}.
\newblock \bibinfo{journal}{\emph{Scientific data}} \bibinfo{volume}{5}, \bibinfo{number}{1} (\bibinfo{year}{2018}), \bibinfo{pages}{1--10}.
\newblock


\bibitem[Lester et~al\mbox{.}(2021)]%
        {prompt}
\bibfield{author}{\bibinfo{person}{Brian Lester}, \bibinfo{person}{Rami Al-Rfou}, {and} \bibinfo{person}{Noah Constant}.} \bibinfo{year}{2021}\natexlab{}.
\newblock \showarticletitle{The power of scale for parameter-efficient prompt tuning}.
\newblock \bibinfo{journal}{\emph{arXiv preprint arXiv:2104.08691}} (\bibinfo{year}{2021}).
\newblock


\bibitem[Li et~al\mbox{.}(2024)]%
        {llava-med}
\bibfield{author}{\bibinfo{person}{Chunyuan Li}, \bibinfo{person}{Cliff Wong}, \bibinfo{person}{Sheng Zhang}, \bibinfo{person}{Naoto Usuyama}, \bibinfo{person}{Haotian Liu}, \bibinfo{person}{Jianwei Yang}, \bibinfo{person}{Tristan Naumann}, \bibinfo{person}{Hoifung Poon}, {and} \bibinfo{person}{Jianfeng Gao}.} \bibinfo{year}{2024}\natexlab{}.
\newblock \showarticletitle{Llava-med: Training a large language-and-vision assistant for biomedicine in one day}.
\newblock \bibinfo{journal}{\emph{Advances in Neural Information Processing Systems}}  \bibinfo{volume}{36} (\bibinfo{year}{2024}).
\newblock


\bibitem[Li et~al\mbox{.}(2023)]%
        {blip2}
\bibfield{author}{\bibinfo{person}{Junnan Li}, \bibinfo{person}{Dongxu Li}, \bibinfo{person}{Silvio Savarese}, {and} \bibinfo{person}{Steven Hoi}.} \bibinfo{year}{2023}\natexlab{}.
\newblock \showarticletitle{Blip-2: Bootstrapping language-image pre-training with frozen image encoders and large language models}. In \bibinfo{booktitle}{\emph{International conference on machine learning}}. PMLR, \bibinfo{pages}{19730--19742}.
\newblock


\bibitem[Liang et~al\mbox{.}(2023)]%
        {xraychat}
\bibfield{author}{\bibinfo{person}{Youwei Liang}, \bibinfo{person}{Han Guo}, {and} \bibinfo{person}{Pengtao Xie}.} \bibinfo{year}{2023}\natexlab{}.
\newblock \showarticletitle{XrayChat: Towards Enabling ChatGPT-Like Capabilities on Chest X-ray Images}.
\newblock  (\bibinfo{year}{2023}).
\newblock


\bibitem[Liang and Li(2024)]%
        {inflora}
\bibfield{author}{\bibinfo{person}{Yan-Shuo Liang} {and} \bibinfo{person}{Wu-Jun Li}.} \bibinfo{year}{2024}\natexlab{}.
\newblock \showarticletitle{InfLoRA: Interference-Free Low-Rank Adaptation for Continual Learning}. In \bibinfo{booktitle}{\emph{Proceedings of the IEEE/CVF Conference on Computer Vision and Pattern Recognition}}. \bibinfo{pages}{23638--23647}.
\newblock


\bibitem[Liu et~al\mbox{.}(2021)]%
        {slake}
\bibfield{author}{\bibinfo{person}{Bo Liu}, \bibinfo{person}{Li-Ming Zhan}, \bibinfo{person}{Li Xu}, \bibinfo{person}{Lin Ma}, \bibinfo{person}{Yan Yang}, {and} \bibinfo{person}{Xiao-Ming Wu}.} \bibinfo{year}{2021}\natexlab{}.
\newblock \showarticletitle{Slake: A semantically-labeled knowledge-enhanced dataset for medical visual question answering}. In \bibinfo{booktitle}{\emph{2021 IEEE 18th International Symposium on Biomedical Imaging (ISBI)}}. IEEE, \bibinfo{pages}{1650--1654}.
\newblock


\bibitem[Liu et~al\mbox{.}(2024)]%
        {llava}
\bibfield{author}{\bibinfo{person}{Haotian Liu}, \bibinfo{person}{Chunyuan Li}, \bibinfo{person}{Qingyang Wu}, {and} \bibinfo{person}{Yong~Jae Lee}.} \bibinfo{year}{2024}\natexlab{}.
\newblock \showarticletitle{Visual instruction tuning}.
\newblock \bibinfo{journal}{\emph{Advances in neural information processing systems}}  \bibinfo{volume}{36} (\bibinfo{year}{2024}).
\newblock


\bibitem[Liu et~al\mbox{.}(2023)]%
        {cl-peft-CIL}
\bibfield{author}{\bibinfo{person}{Xialei Liu}, \bibinfo{person}{Xusheng Cao}, \bibinfo{person}{Haori Lu}, \bibinfo{person}{Jia-wen Xiao}, \bibinfo{person}{Andrew~D Bagdanov}, {and} \bibinfo{person}{Ming-Ming Cheng}.} \bibinfo{year}{2023}\natexlab{}.
\newblock \showarticletitle{Class incremental learning with pre-trained vision-language models}.
\newblock \bibinfo{journal}{\emph{arXiv preprint arXiv:2310.20348}} (\bibinfo{year}{2023}).
\newblock


\bibitem[Lopez-Paz and Ranzato(2017)]%
        {cl-m-gem}
\bibfield{author}{\bibinfo{person}{David Lopez-Paz} {and} \bibinfo{person}{Marc'Aurelio Ranzato}.} \bibinfo{year}{2017}\natexlab{}.
\newblock \showarticletitle{Gradient episodic memory for continual learning}.
\newblock \bibinfo{journal}{\emph{Advances in neural information processing systems}}  \bibinfo{volume}{30} (\bibinfo{year}{2017}).
\newblock


\bibitem[Loshchilov(2017)]%
        {adamw}
\bibfield{author}{\bibinfo{person}{I Loshchilov}.} \bibinfo{year}{2017}\natexlab{}.
\newblock \showarticletitle{Decoupled weight decay regularization}.
\newblock \bibinfo{journal}{\emph{arXiv preprint arXiv:1711.05101}} (\bibinfo{year}{2017}).
\newblock


\bibitem[Mallya and Lazebnik(2018)]%
        {intro-task-il2-packnet}
\bibfield{author}{\bibinfo{person}{Arun Mallya} {and} \bibinfo{person}{Svetlana Lazebnik}.} \bibinfo{year}{2018}\natexlab{}.
\newblock \showarticletitle{Packnet: Adding multiple tasks to a single network by iterative pruning}. In \bibinfo{booktitle}{\emph{Proceedings of the IEEE conference on Computer Vision and Pattern Recognition}}. \bibinfo{pages}{7765--7773}.
\newblock


\bibitem[Oren and Wolf(2021)]%
        {intro-task-il3}
\bibfield{author}{\bibinfo{person}{Guy Oren} {and} \bibinfo{person}{Lior Wolf}.} \bibinfo{year}{2021}\natexlab{}.
\newblock \showarticletitle{In defense of the learning without forgetting for task incremental learning}. In \bibinfo{booktitle}{\emph{Proceedings of the IEEE/CVF International Conference on Computer Vision}}. \bibinfo{pages}{2209--2218}.
\newblock


\bibitem[Papineni et~al\mbox{.}(2002)]%
        {bleu}
\bibfield{author}{\bibinfo{person}{Kishore Papineni}, \bibinfo{person}{Salim Roukos}, \bibinfo{person}{Todd Ward}, {and} \bibinfo{person}{Wei-Jing Zhu}.} \bibinfo{year}{2002}\natexlab{}.
\newblock \showarticletitle{Bleu: a method for automatic evaluation of machine translation}. In \bibinfo{booktitle}{\emph{Proceedings of the 40th annual meeting of the Association for Computational Linguistics}}. \bibinfo{pages}{311--318}.
\newblock


\bibitem[Parisi et~al\mbox{.}(2019)]%
        {intro-cl}
\bibfield{author}{\bibinfo{person}{German~I Parisi}, \bibinfo{person}{Ronald Kemker}, \bibinfo{person}{Jose~L Part}, \bibinfo{person}{Christopher Kanan}, {and} \bibinfo{person}{Stefan Wermter}.} \bibinfo{year}{2019}\natexlab{}.
\newblock \showarticletitle{Continual lifelong learning with neural networks: A review}.
\newblock \bibinfo{journal}{\emph{Neural networks}}  \bibinfo{volume}{113} (\bibinfo{year}{2019}), \bibinfo{pages}{54--71}.
\newblock


\bibitem[Pavlopoulos et~al\mbox{.}(2019)]%
        {iu-x-ray}
\bibfield{author}{\bibinfo{person}{John Pavlopoulos}, \bibinfo{person}{Vasiliki Kougia}, {and} \bibinfo{person}{Ion Androutsopoulos}.} \bibinfo{year}{2019}\natexlab{}.
\newblock \showarticletitle{A survey on biomedical image captioning}. In \bibinfo{booktitle}{\emph{Proceedings of the second workshop on shortcomings in vision and language}}. \bibinfo{pages}{26--36}.
\newblock


\bibitem[Radford et~al\mbox{.}(2021)]%
        {intro-clip}
\bibfield{author}{\bibinfo{person}{Alec Radford}, \bibinfo{person}{Jong~Wook Kim}, \bibinfo{person}{Chris Hallacy}, \bibinfo{person}{Aditya Ramesh}, \bibinfo{person}{Gabriel Goh}, \bibinfo{person}{Sandhini Agarwal}, \bibinfo{person}{Girish Sastry}, \bibinfo{person}{Amanda Askell}, \bibinfo{person}{Pamela Mishkin}, \bibinfo{person}{Jack Clark}, {et~al\mbox{.}}} \bibinfo{year}{2021}\natexlab{}.
\newblock \showarticletitle{Learning transferable visual models from natural language supervision}. In \bibinfo{booktitle}{\emph{International conference on machine learning}}. PMLR, \bibinfo{pages}{8748--8763}.
\newblock


\bibitem[Rebuffi et~al\mbox{.}(2017)]%
        {cl-m-icarl}
\bibfield{author}{\bibinfo{person}{Sylvestre-Alvise Rebuffi}, \bibinfo{person}{Alexander Kolesnikov}, \bibinfo{person}{Georg Sperl}, {and} \bibinfo{person}{Christoph~H Lampert}.} \bibinfo{year}{2017}\natexlab{}.
\newblock \showarticletitle{icarl: Incremental classifier and representation learning}. In \bibinfo{booktitle}{\emph{Proceedings of the IEEE conference on Computer Vision and Pattern Recognition}}. \bibinfo{pages}{2001--2010}.
\newblock


\bibitem[Saha et~al\mbox{.}(2021)]%
        {cl-r-GPM}
\bibfield{author}{\bibinfo{person}{Gobinda Saha}, \bibinfo{person}{Isha Garg}, {and} \bibinfo{person}{Kaushik Roy}.} \bibinfo{year}{2021}\natexlab{}.
\newblock \showarticletitle{Gradient Projection Memory for Continual Learning}. In \bibinfo{booktitle}{\emph{International Conference on Learning Representations}}.
\newblock
\urldef\tempurl%
\url{https://openreview.net/forum?id=3AOj0RCNC2}
\showURL{%
\tempurl}


\bibitem[Seyfioglu et~al\mbox{.}(2024)]%
        {quilt-llava}
\bibfield{author}{\bibinfo{person}{Mehmet~Saygin Seyfioglu}, \bibinfo{person}{Wisdom~O Ikezogwo}, \bibinfo{person}{Fatemeh Ghezloo}, \bibinfo{person}{Ranjay Krishna}, {and} \bibinfo{person}{Linda Shapiro}.} \bibinfo{year}{2024}\natexlab{}.
\newblock \showarticletitle{Quilt-llava: Visual instruction tuning by extracting localized narratives from open-source histopathology videos}. In \bibinfo{booktitle}{\emph{Proceedings of the IEEE/CVF Conference on Computer Vision and Pattern Recognition}}. \bibinfo{pages}{13183--13192}.
\newblock


\bibitem[Soutif-Cormerais et~al\mbox{.}(2021)]%
        {intro-task-il}
\bibfield{author}{\bibinfo{person}{Albin Soutif-Cormerais}, \bibinfo{person}{Marc Masana}, \bibinfo{person}{Joost Van~de Weijer}, {and} \bibinfo{person}{Bartl{\o}miej Twardowski}.} \bibinfo{year}{2021}\natexlab{}.
\newblock \showarticletitle{On the importance of cross-task features for class-incremental learning}.
\newblock \bibinfo{journal}{\emph{arXiv preprint arXiv:2106.11930}}  \bibinfo{volume}{1} (\bibinfo{year}{2021}).
\newblock


\bibitem[Sun et~al\mbox{.}(2023)]%
        {pathasst}
\bibfield{author}{\bibinfo{person}{Yuxuan Sun}, \bibinfo{person}{Chenglu Zhu}, \bibinfo{person}{Sunyi Zheng}, \bibinfo{person}{Kai Zhang}, \bibinfo{person}{Zhongyi Shui}, \bibinfo{person}{Xiaoxuan Yu}, \bibinfo{person}{Yizhi Zhao}, \bibinfo{person}{Honglin Li}, \bibinfo{person}{Yunlong Zhang}, \bibinfo{person}{Ruojia Zhao}, {et~al\mbox{.}}} \bibinfo{year}{2023}\natexlab{}.
\newblock \showarticletitle{Pathasst: Redefining pathology through generative foundation ai assistant for pathology}.
\newblock \bibinfo{journal}{\emph{arXiv preprint arXiv:2305.15072}} (\bibinfo{year}{2023}).
\newblock


\bibitem[Sung et~al\mbox{.}(2022)]%
        {intro-preft3}
\bibfield{author}{\bibinfo{person}{Yi-Lin Sung}, \bibinfo{person}{Jaemin Cho}, {and} \bibinfo{person}{Mohit Bansal}.} \bibinfo{year}{2022}\natexlab{}.
\newblock \showarticletitle{Vl-adapter: Parameter-efficient transfer learning for vision-and-language tasks}. In \bibinfo{booktitle}{\emph{Proceedings of the IEEE/CVF conference on computer vision and pattern recognition}}. \bibinfo{pages}{5227--5237}.
\newblock


\bibitem[Tschandl et~al\mbox{.}(2018)]%
        {HAM}
\bibfield{author}{\bibinfo{person}{Philipp Tschandl}, \bibinfo{person}{Cliff Rosendahl}, {and} \bibinfo{person}{Harald Kittler}.} \bibinfo{year}{2018}\natexlab{}.
\newblock \showarticletitle{The HAM10000 dataset, a large collection of multi-source dermatoscopic images of common pigmented skin lesions. Scientific Data. 2018; 5: 180161}.
\newblock \bibinfo{journal}{\emph{Search in}}  \bibinfo{volume}{2} (\bibinfo{year}{2018}).
\newblock


\bibitem[Tu et~al\mbox{.}(2024)]%
        {GBAI}
\bibfield{author}{\bibinfo{person}{Tao Tu}, \bibinfo{person}{Shekoofeh Azizi}, \bibinfo{person}{Danny Driess}, \bibinfo{person}{Mike Schaekermann}, \bibinfo{person}{Mohamed Amin}, \bibinfo{person}{Pi-Chuan Chang}, \bibinfo{person}{Andrew Carroll}, \bibinfo{person}{Charles Lau}, \bibinfo{person}{Ryutaro Tanno}, \bibinfo{person}{Ira Ktena}, {et~al\mbox{.}}} \bibinfo{year}{2024}\natexlab{}.
\newblock \showarticletitle{Towards generalist biomedical AI}.
\newblock \bibinfo{journal}{\emph{NEJM AI}} \bibinfo{volume}{1}, \bibinfo{number}{3} (\bibinfo{year}{2024}), \bibinfo{pages}{AIoa2300138}.
\newblock


\bibitem[Veeling et~al\mbox{.}(2018)]%
        {pcam}
\bibfield{author}{\bibinfo{person}{Bastiaan~S Veeling}, \bibinfo{person}{Jasper Linmans}, \bibinfo{person}{Jim Winkens}, \bibinfo{person}{Taco Cohen}, {and} \bibinfo{person}{Max Welling}.} \bibinfo{year}{2018}\natexlab{}.
\newblock \showarticletitle{Rotation equivariant CNNs for digital pathology}. In \bibinfo{booktitle}{\emph{Medical Image Computing and Computer Assisted Intervention--MICCAI 2018: 21st International Conference, Granada, Spain, September 16-20, 2018, Proceedings, Part II 11}}. Springer, \bibinfo{pages}{210--218}.
\newblock


\bibitem[Wang et~al\mbox{.}(2024)]%
        {intro-survey-cl}
\bibfield{author}{\bibinfo{person}{Liyuan Wang}, \bibinfo{person}{Xingxing Zhang}, \bibinfo{person}{Hang Su}, {and} \bibinfo{person}{Jun Zhu}.} \bibinfo{year}{2024}\natexlab{}.
\newblock \showarticletitle{A comprehensive survey of continual learning: theory, method and application}.
\newblock \bibinfo{journal}{\emph{IEEE Transactions on Pattern Analysis and Machine Intelligence}} (\bibinfo{year}{2024}).
\newblock


\bibitem[Wu et~al\mbox{.}({[n.\,d.]})]%
        {sd-lora}
\bibfield{author}{\bibinfo{person}{Yichen Wu}, \bibinfo{person}{Hongming Piao}, \bibinfo{person}{Long-Kai Huang}, \bibinfo{person}{Renzhen Wang}, \bibinfo{person}{Wanhua Li}, \bibinfo{person}{Hanspeter Pfister}, \bibinfo{person}{Deyu Meng}, \bibinfo{person}{Kede Ma}, {and} \bibinfo{person}{Ying Wei}.} \bibinfo{year}{[n.\,d.]}\natexlab{}.
\newblock \showarticletitle{SD-LoRA: Scalable Decoupled Low-Rank Adaptation for Class Incremental Learning}. In \bibinfo{booktitle}{\emph{The Thirteenth International Conference on Learning Representations}}.
\newblock


\bibitem[Yu et~al\mbox{.}(2024)]%
        {cl-peft-ddas}
\bibfield{author}{\bibinfo{person}{Jiazuo Yu}, \bibinfo{person}{Yunzhi Zhuge}, \bibinfo{person}{Lu Zhang}, \bibinfo{person}{Ping Hu}, \bibinfo{person}{Dong Wang}, \bibinfo{person}{Huchuan Lu}, {and} \bibinfo{person}{You He}.} \bibinfo{year}{2024}\natexlab{}.
\newblock \showarticletitle{Boosting continual learning of vision-language models via mixture-of-experts adapters}. In \bibinfo{booktitle}{\emph{Proceedings of the IEEE/CVF Conference on Computer Vision and Pattern Recognition}}. \bibinfo{pages}{23219--23230}.
\newblock


\bibitem[Zenke et~al\mbox{.}(2017)]%
        {cl-r-is}
\bibfield{author}{\bibinfo{person}{Friedemann Zenke}, \bibinfo{person}{Ben Poole}, {and} \bibinfo{person}{Surya Ganguli}.} \bibinfo{year}{2017}\natexlab{}.
\newblock \showarticletitle{Continual learning through synaptic intelligence}. In \bibinfo{booktitle}{\emph{International conference on machine learning}}. PMLR, \bibinfo{pages}{3987--3995}.
\newblock


\bibitem[Zhang et~al\mbox{.}(2024)]%
        {BiomedGPT}
\bibfield{author}{\bibinfo{person}{Kai Zhang}, \bibinfo{person}{Rong Zhou}, \bibinfo{person}{Eashan Adhikarla}, \bibinfo{person}{Zhiling Yan}, \bibinfo{person}{Yixin Liu}, \bibinfo{person}{Jun Yu}, \bibinfo{person}{Zhengliang Liu}, \bibinfo{person}{Xun Chen}, \bibinfo{person}{Brian~D Davison}, \bibinfo{person}{Hui Ren}, {et~al\mbox{.}}} \bibinfo{year}{2024}\natexlab{}.
\newblock \showarticletitle{A generalist vision--language foundation model for diverse biomedical tasks}.
\newblock \bibinfo{journal}{\emph{Nature Medicine}} (\bibinfo{year}{2024}), \bibinfo{pages}{1--13}.
\newblock


\bibitem[Zhang et~al\mbox{.}(2023)]%
        {adalora}
\bibfield{author}{\bibinfo{person}{Qingru Zhang}, \bibinfo{person}{Minshuo Chen}, \bibinfo{person}{Alexander Bukharin}, \bibinfo{person}{Nikos Karampatziakis}, \bibinfo{person}{Pengcheng He}, \bibinfo{person}{Yu Cheng}, \bibinfo{person}{Weizhu Chen}, {and} \bibinfo{person}{Tuo Zhao}.} \bibinfo{year}{2023}\natexlab{}.
\newblock \showarticletitle{AdaLoRA: Adaptive budget allocation for parameter-efficient fine-tuning}.
\newblock \bibinfo{journal}{\emph{arXiv preprint arXiv:2303.10512}} (\bibinfo{year}{2023}).
\newblock


\bibitem[Zhang et~al\mbox{.}(2022)]%
        {intro-peft4-tip-adapter}
\bibfield{author}{\bibinfo{person}{Renrui Zhang}, \bibinfo{person}{Wei Zhang}, \bibinfo{person}{Rongyao Fang}, \bibinfo{person}{Peng Gao}, \bibinfo{person}{Kunchang Li}, \bibinfo{person}{Jifeng Dai}, \bibinfo{person}{Yu Qiao}, {and} \bibinfo{person}{Hongsheng Li}.} \bibinfo{year}{2022}\natexlab{}.
\newblock \showarticletitle{Tip-adapter: Training-free adaption of clip for few-shot classification}. In \bibinfo{booktitle}{\emph{European conference on computer vision}}. Springer, \bibinfo{pages}{493--510}.
\newblock


\bibitem[Zhang et~al\mbox{.}(2019)]%
        {wsi-diagnosis}
\bibfield{author}{\bibinfo{person}{Zizhao Zhang}, \bibinfo{person}{Pingjun Chen}, \bibinfo{person}{Mason McGough}, \bibinfo{person}{Fuyong Xing}, \bibinfo{person}{Chunbao Wang}, \bibinfo{person}{Marilyn Bui}, \bibinfo{person}{Yuanpu Xie}, \bibinfo{person}{Manish Sapkota}, \bibinfo{person}{Lei Cui}, \bibinfo{person}{Jasreman Dhillon}, {et~al\mbox{.}}} \bibinfo{year}{2019}\natexlab{}.
\newblock \showarticletitle{Pathologist-level interpretable whole-slide cancer diagnosis with deep learning}.
\newblock \bibinfo{journal}{\emph{Nature Machine Intelligence}} \bibinfo{volume}{1}, \bibinfo{number}{5} (\bibinfo{year}{2019}), \bibinfo{pages}{236}.
\newblock


\bibitem[Zhou et~al\mbox{.}(2023)]%
        {skingpt}
\bibfield{author}{\bibinfo{person}{Juexiao Zhou}, \bibinfo{person}{Xiaonan He}, \bibinfo{person}{Liyuan Sun}, \bibinfo{person}{Jiannan Xu}, \bibinfo{person}{Xiuying Chen}, \bibinfo{person}{Yuetan Chu}, \bibinfo{person}{Longxi Zhou}, \bibinfo{person}{Xingyu Liao}, \bibinfo{person}{Bin Zhang}, {and} \bibinfo{person}{Xin Gao}.} \bibinfo{year}{2023}\natexlab{}.
\newblock \showarticletitle{SkinGPT-4: an interactive dermatology diagnostic system with visual large language model}.
\newblock  (\bibinfo{year}{2023}).
\newblock


\bibitem[Zhou et~al\mbox{.}(2022b)]%
        {cocoop}
\bibfield{author}{\bibinfo{person}{Kaiyang Zhou}, \bibinfo{person}{Jingkang Yang}, \bibinfo{person}{Chen~Change Loy}, {and} \bibinfo{person}{Ziwei Liu}.} \bibinfo{year}{2022}\natexlab{b}.
\newblock \showarticletitle{Conditional prompt learning for vision-language models}. In \bibinfo{booktitle}{\emph{Proceedings of the IEEE/CVF conference on computer vision and pattern recognition}}. \bibinfo{pages}{16816--16825}.
\newblock


\bibitem[Zhou et~al\mbox{.}(2022c)]%
        {coop}
\bibfield{author}{\bibinfo{person}{Kaiyang Zhou}, \bibinfo{person}{Jingkang Yang}, \bibinfo{person}{Chen~Change Loy}, {and} \bibinfo{person}{Ziwei Liu}.} \bibinfo{year}{2022}\natexlab{c}.
\newblock \showarticletitle{Learning to prompt for vision-language models}.
\newblock \bibinfo{journal}{\emph{International Journal of Computer Vision}} \bibinfo{volume}{130}, \bibinfo{number}{9} (\bibinfo{year}{2022}), \bibinfo{pages}{2337--2348}.
\newblock


\bibitem[Zhou et~al\mbox{.}(2022a)]%
        {intro-moe}
\bibfield{author}{\bibinfo{person}{Yanqi Zhou}, \bibinfo{person}{Tao Lei}, \bibinfo{person}{Hanxiao Liu}, \bibinfo{person}{Nan Du}, \bibinfo{person}{Yanping Huang}, \bibinfo{person}{Vincent Zhao}, \bibinfo{person}{Andrew~M Dai}, \bibinfo{person}{Quoc~V Le}, \bibinfo{person}{James Laudon}, {et~al\mbox{.}}} \bibinfo{year}{2022}\natexlab{a}.
\newblock \showarticletitle{Mixture-of-experts with expert choice routing}.
\newblock \bibinfo{journal}{\emph{Advances in Neural Information Processing Systems}}  \bibinfo{volume}{35} (\bibinfo{year}{2022}), \bibinfo{pages}{7103--7114}.
\newblock


\bibitem[Zhu et~al\mbox{.}(2023)]%
        {minigpt4}
\bibfield{author}{\bibinfo{person}{Deyao Zhu}, \bibinfo{person}{Jun Chen}, \bibinfo{person}{Xiaoqian Shen}, \bibinfo{person}{Xiang Li}, {and} \bibinfo{person}{Mohamed Elhoseiny}.} \bibinfo{year}{2023}\natexlab{}.
\newblock \showarticletitle{Minigpt-4: Enhancing vision-language understanding with advanced large language models}.
\newblock \bibinfo{journal}{\emph{arXiv preprint arXiv:2304.10592}} (\bibinfo{year}{2023}).
\newblock


\end{thebibliography}

\end{document}